\documentclass[lettersize,journal]{IEEEtran}
\usepackage{amsmath,amsfonts}
\usepackage{algorithmic}
\usepackage{algorithm}
\usepackage{array}
\usepackage[caption=false,font=normalsize,labelfont=sf,textfont=sf]{subfig}
\usepackage{textcomp}
\usepackage{stfloats}
\usepackage{url}
\usepackage{verbatim}
\usepackage{graphicx}
\usepackage{cite}
\usepackage{amssymb, booktabs}
\usepackage{epsfig}
\usepackage{multirow}
\usepackage[pagebackref=true,breaklinks=true,letterpaper=true,colorlinks,citecolor=blue,linkcolor=blue,bookmarks=false]{hyperref}
\usepackage[T1]{fontenc}
\usepackage{tikz,xcolor}

\definecolor{lime}{HTML}{A6CE39}
\DeclareRobustCommand{\orcidicon}{
	\begin{tikzpicture}
		\draw[lime, fill=lime] (0,0)
		circle[radius=0.16]
		node[white]{{\fontfamily{qag}\selectfont \tiny \.{I}D}};
	\end{tikzpicture}
	\hspace{-2mm}
}
\foreach \x in {A, ..., Z}{%
	\expandafter\xdef\csname orcid\x\endcsname{\noexpand\href{https://orcid.org/\csname orcidauthor\x\endcsname}{\noexpand\orcidicon}}
}

\begin{document}

\title{Multi-Exposure Image Fusion via Distilled 3D LUT Grid with Editable Mode}

\author{Xin Su$^{1}$,~and~Zhuoran Zheng$^{2*}$,~\IEEEmembership{Member,~IEEE} \\
$^{1}$ Fuzhou University, $^{2}$ Sun Yat-sen University}



\maketitle

\begin{abstract}
With the rising imaging resolution of handheld devices, existing multi-exposure image fusion algorithms struggle to generate a high dynamic range image with ultra-high resolution in real-time.
Apart from that, there is a trend to design a manageable and editable algorithm as the different needs of real application scenarios.
To tackle these issues, we introduce 3D LUT technology, which can enhance images with ultra-high-definition (UHD) resolution in real time on resource-constrained devices.
However, since the fusion of information from multiple images with different exposure rates is uncertain, and this uncertainty significantly trials the generalization power of the 3D LUT grid.
To address this issue and ensure a robust learning space for the model, we propose using a teacher-student network to model the uncertainty on the 3D LUT grid.
Furthermore, we provide an editable mode for the multi-exposure image fusion algorithm by using the implicit representation function to match the requirements in different scenarios.
Extensive experiments demonstrate that our proposed method is highly competitive in efficiency and accuracy.
%
%
The code is released at \url{https://github.com/zzr-idam/UHD-multi-exposure-image-fusion-algorithm}.
\end{abstract}


\section{Introduction} \label{sec:intro}
To boost the dynamic range of natural images recorded by handheld devices, \textbf{M}ulti-\textbf{E}xposure image \textbf{F}usion (MEF) techniques~\cite{li2018multi,qu2022transmef,ram2017deepfuse,xu2020fusiondn,xu2020mef,zhang2020ifcnn,yu2022efficient,yan2022lightweight,li2022gamma,Zheng_2023_CVPR,luo2023multi} are widely employed to generate a visually pleasing image.
Recently, with the continuous development and improvement of hardware technology, photos or videos from handheld devices can usually be recorded with ultra-high-definition (UHD) resolution.
Despite the advantages of UHD resolution, the high density of pixels (in the millions) can cause existing Multi-Exposure Fusion (MEF) algorithms to fail when running on resource-limited devices. 
This is particularly true for deep learning approaches such as a simple CNN with three layers of convolution, where fusion of more than 3 UHD images with differing exposures at once could cause the CPU on a mobile phone to be unable to handle the workload.
Apart from the aforementioned limitation, current MEF algorithms also lack customization options to cater to unique demands of different scenarios.
\begin{figure}[t]
	\centering
	\includegraphics[width=0.495\textwidth]{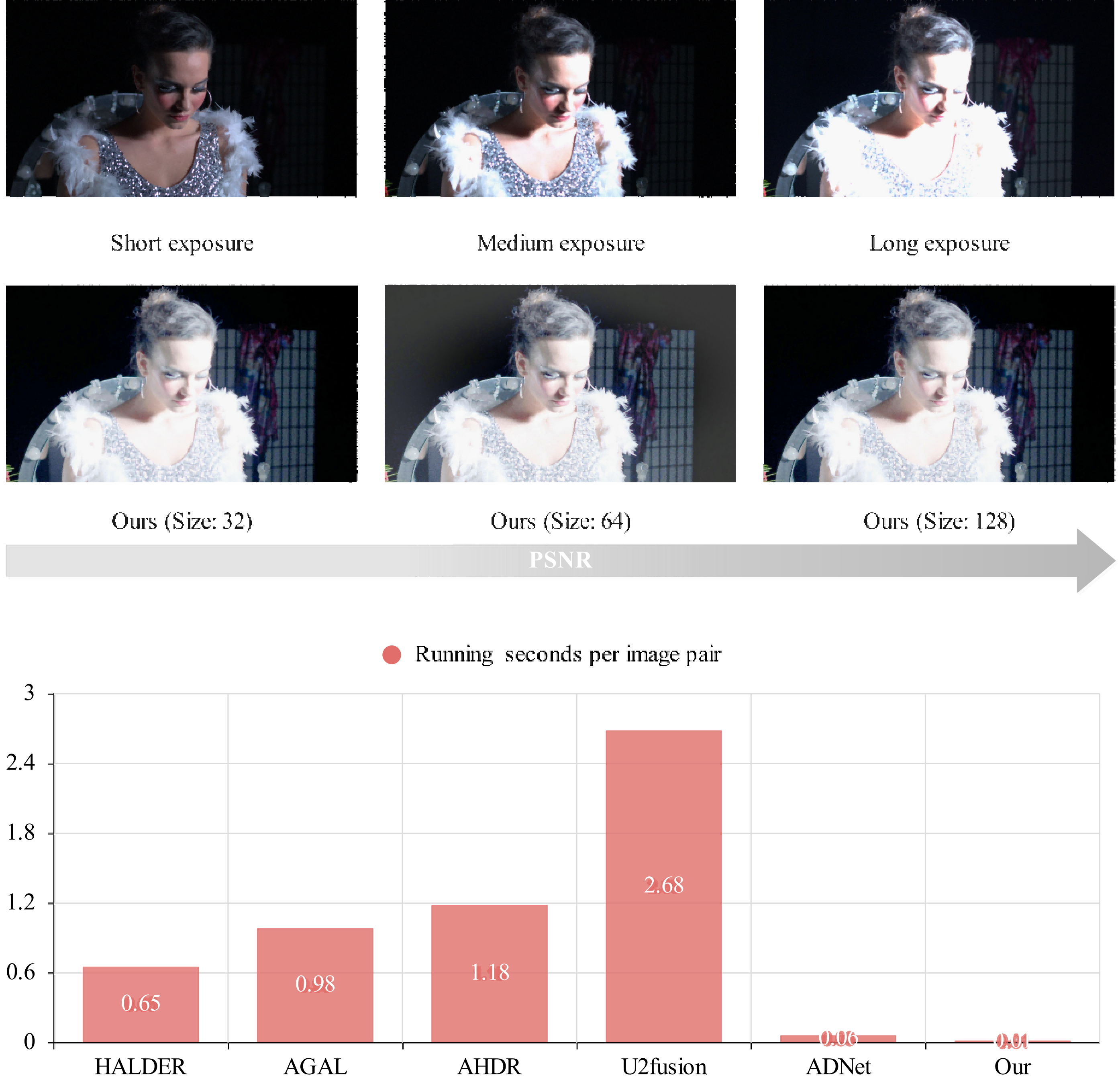}
	\vspace{-1mm}
	\caption{
		The top figure shows the results of our method run on a multi-exposure dataset.
		\textbf{Size} illustrates the scale of the 3D LUT grid, for example, 32 means the size of the grid is 32 $\times$ 32 $\times$ 32, and the PSNR is gradually increased with the size of the grid.
		Note that the scale of this grid is editable due to our use of implicit neural representation.
		The bottom figure shows the average running time comparison over MEFB, a dataset
		containing 50 image pairs of average size $3 \times 551 \times 707$.
	}
	\vspace{-4mm}
	\label{f1}
\end{figure}

Faced with the limitations of these MEF algorithms, our research explores general techniques for enhancing UHD images.
So far, there are three mainstream deep learning-based solutions: 1) Bilateral learning~\cite{ZhengRCWJ21}; 2) 3D LUT~\cite{ZengCLCZ22}; and 3) Laplace pyramid~\cite{lap}.
All of these methods are designed to speed up the process of enhancing UHD images with different resolutions. Notably, the 3D LUT-based algorithm can process a 4K image in just 9ms on a single GPU. This is made possible by incorporating prior knowledge of image retouching techniques. The algorithm pre-designs several toned grids and uses a deep network to learn the weights of these grids for fusion. Finally, the fused grids are applied to the raw image for enhancement.
To trade-off the accuracy and efficiency of the regression, here we select a 3D LUT-based deep learning method as a base to generate a high dynamic range (HDR) image with UHD resolution.
Although 3D LUT has the potential to handle UHD MEF tasks, existing 3D LUT techniques only enhance a single input image, and we need to address the following concerns when faced with a set of the input image.
\textbf{i)} How to generate a robust grid acting on the raw information when faced with multiple images whose luminance, texture, and color are different (facing the issue of information fusion with \textbf{uncertainty}).
\textbf{ii)} In addition, on which \textbf{raw information} (synthetic information or an raw image) should the grid act?
\textbf{iii)} How to extend this technology into a customized algorithm that is manageable and editable?
\begin{figure*}[t]
	\centering
	\includegraphics[width=0.910\textwidth]{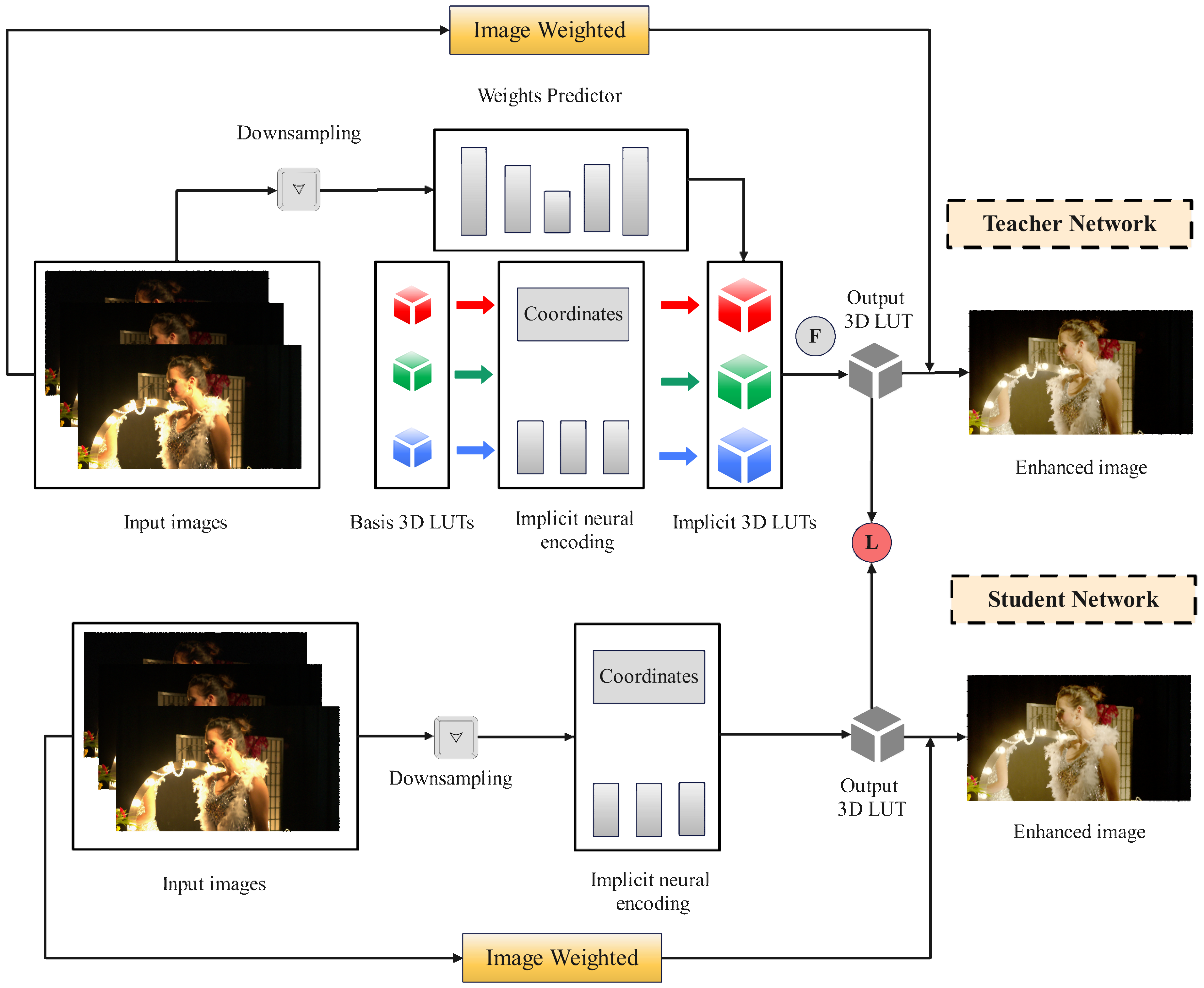}
	\vspace{-1mm}
	\caption{
		\textbf{The architecture of our approach}.
		This figure shows a learning paradigm for a student-teacher network. First, the teacher network learns a high-quality 3D LUT, after that, the 3D LUT in the student network is constrained by the teacher network, and finally, the student network generates a robust 3D LUT.
		\textbf{F} denotes the weighted fusion strategy and \textbf{L} denotes the restricted loss term.
	}
	\vspace{-2mm}
	\label{frameworks}
\end{figure*}

To tackle these issues, we develop a method that leverages a teacher-student network to build a distilled 3D LUT grid for fusing a set of low dynamic range images~\cite{YeB22}. 
Specifically, we constrain the 3D LUT grid to model the uncertainty of the input information by encoding the correlation between the patch of true image and then selecting the raw images to be enhanced with the help of a classifier (a small-scale \texttt{CNN} with \texttt{MLP}).
The whole 3D LUT encoder is modeled by an implicit neural network (student network)~\cite{NguyenB23}, which is an editable pattern to generate high dynamic range images of varying quality.
Our approach employs a lightweight network that can process 5 UHD resolution images on a single GPU, achieving real-time processing speed of at least 33fps. In our experimental section, we evaluate the algorithm using several publicly available datasets as well as a customized dataset. Results from a large number of experiments demonstrate the effectiveness of our approach.
The contributions of this paper are summarized as follows:
\begin{itemize}
	\vspace{0mm}
	\item We propose a teacher-student network architecture for learning to a distilled 3D LUT grid, which boosts the generalization ability of the student network by modeling uncertainty.
	\vspace{0mm}
	\item We introduce an implicit neural network to build a student network to yield a 3D LUT grid of arbitrary size, which can be used to generate UHD images of different quality based on the scene conditions.
	\vspace{0mm}
	\item By discussing and experimenting to bridge the gap between the three popular UHD degradation image enhancement methods, in addition, a large number of experiments (quantitative and qualitative evaluations) demonstrate the effectiveness of our approach.
\end{itemize}

\section{Related Work}

\noindent \textbf{Image fusion methods.}
Conventional methods for image fusion include spatial domain-based methods and transformation domain-based algorithms. Spatial domain approaches~\cite{lee2018multi,ma2017multi,ma2015multi} analyze the information significance of raw images and fuse them spatially using an estimated weight map. On the other hand, transformation domain approaches~\cite{li2020fast} concentrate on the coefficients of decomposed basis vectors and assess the importance of these signals by using a simple model before fusion.

Recently, deep learning-based approaches have delivered promising results in the MEF field.
Based on the convolutional neural network (CNN), Deepfuse~\cite{ram2017deepfuse} first proposes that the merging of luminance maps is represented by deep learning and fuses chromaticity different parts by conventional weighted averaging methods.
Subsequently, a large number of CNN-based approaches are proposed, such as those based on generative adversarial networks~\cite{mei2018progressive}.
However, these methods are a local modeling strategy and they struggle to capture the global information of the image, and to solve this problem, Transformer-based methods are proposed.
Although these methods are demonstrated to be effective on MEF tasks, they usually require stacking a large number of convolutional layers and attention modules, which can yield artifacts or ghost motifs easily in real scenes.
In addition, unsupervised and self-supervised based methods have been heavily introduced to alleviate the problems of unpaired images and over-fitting.
Most of the above models cannot directly process 4K or higher resolution images on a single GPU shader with 24G RAM. 

\noindent \textbf{LUTs for image enhancement.}
3D look-up tables (LUTs) are ultra-fast algorithms for color mapping and are widely used to improve the quality of digital images for color correction, video enhancement, and retouching.
In recent years, many neural network-based LUT generation methods are proposed~\cite{karaimer2016software,yang2022seplut,zeng2020learning,zhang2022clut}. 
Image-Adaptive 3DLUT uses a simple CNN weight predictor to estimate several basic 3D LUT weights learned. 
Then, an adaptive 3D LUT is generated for each input image based on the image content fusion 3D LUT.
Afterwards, the 3D LUTs are fused to generate a grid containing the affine transform coefficients, and finally this grid acts on the raw image to achieve image enhancement.
4D LUT proposes a context-aware 4D lookup table that can be adapted to learn the photo context to achieve content-dependent enhancement of different contents in each image. 
CLUT analyses the inherent compressibility of 3D LUT and proposes an efficient 3D LUT compression representation that maintains the powerful mapping capability of 3D LUT.
These methods explore the use of neural networks to generate LUTs, showing the robust capabilities of LUTs in image and video processing.
%
%
%

\noindent \textbf{UHD image processing methods.}
To transfer our laboratory models to real application scenarios effectively, HD image research\cite{he2021inferring,song2021starenhancer,verelst2021blockcopy,wang2019underexposed,wang2021real,wu2021contrastive,yuan2021hrformer,zhang2021multi} which can be processed in real-time is becoming popular. 
After that, with the development of deep learning techniques, some methods for real-time processing of UHD images are proposed.
However, at present, the method for UHD multi-exposure image processing is still in a limbo stage.

\begin{figure}[t]
	\centering
	\includegraphics[width=0.452\textwidth]{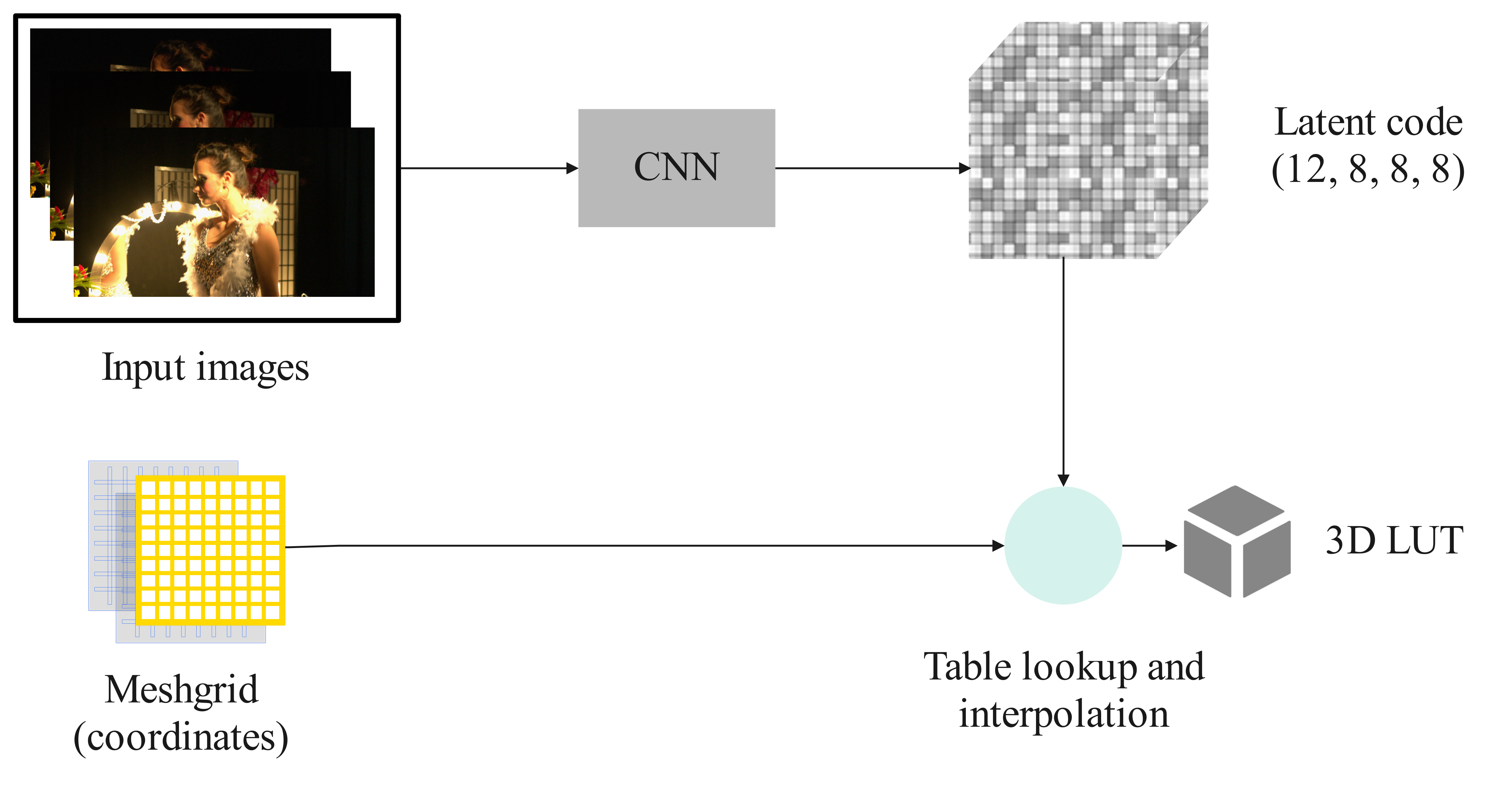}
	\caption{
		\textbf{The architecture of implicit neural network}.
	}
	\vspace{-4mm}
	\label{f2}
\end{figure}

\section{Method}
The workflow of our approach is illustrated in Figure~\ref{frameworks} (best viewed in color).
Through a distillation loss $\mathcal{L_\text{grid}}$ bridges the teacher network and the student network; it can be noticed that the student network does not with customized 3D LUTs, rather it just regresses a cubic grid using an implicit neural network.
Our approach is depicted by five sub-summaries.

\noindent \textbf{Generating a 3D lookup tables.}
In this paper, we view a 3D LUT as an implicit neural network regression target (student network), which is a continuous space where the size of the grid is in \textbf{editable mode}.
In a nutshell, an implicit neural network $\mathbf{F}_{\texttt{INN}}$ receives a pair of information (latent code $l$ and coordinates $c$) that can be regressed to the element $e$ corresponding to that coordinate value.
Here the latent code $l$ is a feature map by a lightweight CNN encoding the input information, here is a feature map of size $12 \times 8 \times 8 \times 8$.
The coordinates $c$ are in editable mode and the dimensions can be generated arbitrarily.
The whole flow is shown in Figure~\ref{f2}, where the table lookup and interpolation use \url{https://pytorch.org/docs/stable/generated/torch.nn.functional.grid_sample.html#torch.nn.functional.grid_sample}.
This process can be written as
\begin{equation}
	v_{(i,j,k)} = \mathbf{F}_{\texttt{INN}}(l, c(i,j,k)),
\end{equation}
where $i,j,k$ denotes the coordinates on the three-dimensional space, $v$ denotes the element at each position in the 3D LUT $V$.
Finally, the 3D LUT grid $V$ acts on a given raw information (qualified by a network) to generate an enhanced image.
For the teacher network, to fuse a high-quality grid $\hat{V}$ needs to input three prior grids ($V1 = \{v1_{r},v1_{g},v1_{b},V2 = \{v2_{r},v2_{g},v2_{b},V3 = \{v3_{r},v3_{g},v3_{b}\}$) into a simple \texttt{CNN} to generate three $\{L_{1},L_{2},L_{3},\}$; after that, they are input to the implicit neural network $\mathbf{F}_{\texttt{INFF}}$ to obtain three enhanced grids ($\hat{V}_{1}, \hat{V}_{2},\hat{V}_{3}$); finally, they are predicted by \texttt{Weights Predictor} (WP) to conduct a linear fusion of the three weights ($w_{1}, w_{2}, w_{3}$).
WP is a lightweight CNN that predicts a set of weights from the input images.

\noindent \textbf{Teacher network.}
The teacher network can be viewed as a standard 3D LUT model, where the only concern is simply the information input from multiple images $I_{*}$.
Specifically, the primary issue to consider is computational efficiency. 
To obtain fast inference capability, the extant methods for handling UHD images require large-scale downsampling of the raw input (3840 $\times$ 2160 $\rightarrow$ 256 $\times$ 256).
Here, we downsample the resolution of the raw images to 128 $\times$ 128.
Then, the $I_{*}$'s are stacked in the channel dimension $\mathbb{C}$, following a sequence of gradually growing exposure.
The input information that is pre-processed then passes to two networks (IW [Image Weighted], WP [Weights Predictor]).
WP aims to provide weights for fusing 3D LUTs $V$.
WP is a U-Net network~\cite{miccais} with several layers of \texttt{MLPs}, and the number of neurons in the last layer of \texttt{MLPs} corresponds to the number of basis 3D LUTs.
\begin{equation}
	w_{*} = \text{WP}(\text{S}(I_{*})),
\end{equation}
where S denotes that the images are stacked on the channel domain $\mathbb{C}$.
The IW aims to provide a high-quality reconstruction target for $\hat{V}$, which feeds the input images with the weights required for fusion.
IW utilizes only several 3 $\times$ 3 convolutions and two layers of \texttt{MLP}, with the last layer being \texttt{Softmax}.
This algorithm can be formalized as follows
\begin{equation}
	\text{I}_{f} = \text{IW}(\text{S}(I_{*})) \cdot \text{I},
\end{equation}
where $\text{I}_{f}$ represents the fused image and I represent a set of multi-exposure images with the original resolution.
In general, the whole inference process of the teacher network can be written 
\begin{equation}
	\text{I}_{E} = \hat{V}~~\Delta~~\text{I}_{f},
\end{equation}
where $\text{I}_{E}$ represents the obtained UHD-enhanced image and $\Delta$ represents trilinear interpolation and lookup scheme.

\noindent \textbf{Student network.}
The student network is a \textcolor{blue}{lightweight network} that does not rely on any prior knowledge and regresses a 3D LUT grid through a set of networks (IW and implicit neural network).
The overall workflow can be written
\begin{equation}
	\text{I}_{E} = V~~\Delta~~\text{I}_{f},
\end{equation}

\noindent \textbf{Loss functions.}
The key to our approach lies in the design of the loss function.
For the teacher network, we use only one $\mathbf{L}_{1}$ loss, which focuses on the difference between each pixel value.
For the student network, we need to trade off the three loss function terms to generate a robust 3D LUT grid $V$.
Specifically, the first is the $\mathbf{L}_{1}$ loss function, which computes the distance between $\text{I}_{E}$ (output of the algorithm) and the truth image $\text{I}_{G}$.
The second loss term $\mathcal{L}_{d1}$ is the distillation loss about grid $V$. 
The distance between $V \in \mathbb{R}^{64 \times 64 \times 64}$ and $\hat{V} \in \mathbb{R}^{64 \times 64 \times 64}$ is measured through an $\mathbf{L}_{1}$ loss.
The third loss term $\mathcal{L}_{lr}$ aims to boost the robustness of the student network, and it has the assumption that the long-range dependencies between the elements of the grid should be similar to $\text{I}_{G}$.
Specifically, $\text{I}_{G}$ is tokenized (similar to \texttt{ViT}~\cite{vit}'s serialization step) and then gets a correlation matrix $\mathcal{M}_g \in \mathbb{R}^{16 \times 16}$ between image patches by the dot product.
$V$ is tokenized through a small \texttt{CNN} network to obtain a matrix $\mathcal{M}_b \in \mathbb{R}^{16 \times 16}$.
The assumption of $\mathcal{L}_{lr}$ is based on our observations of the data distribution of each of $V$ and $\text{I}_{G}$.
This assumption is explored in the discussion section. 
The total loss function $\mathcal{L}_{\text{total}}$ can be written
\begin{equation}
	\mathcal{L}_{\text{total}} = \mathcal{L}_{1} + \alpha \mathcal{L}_{d1} + \beta \mathcal{L}_{lr},
\end{equation}
where $\alpha$ and $\beta$ were set to 0.05 and 0.09, respectively; in addition, we also try to enforce a perceptual loss~\cite{ploss} on the student network, but the visual effect is not significant.

\noindent \textbf{Details of the neural network.}
For the teacher network, the implicit neural coding comes with two sub-networks, where the \texttt{CNN} is a 5+2 network (5 convolutional layers with 3 $\times$ 3 convolutional kernels, the activation function uses \texttt{ReLU}); the \texttt{MLP} has only 5 layers (the activation function uses \texttt{ELU}).
For the student network, the implicit neural coding comes with two sub-networks, where the \texttt{CNN} is an 8+2 network (8 convolutional layers with 3 $\times$ 3 convolutional kernels, the activation function uses \texttt{ReLU}); the \texttt{MLP} has only 7 layers (the activation function uses \texttt{ELU}).
IW employs a 3-layer convolution with a 2-layer \texttt{MLP}, and the activation function uses \texttt{ReLU}.
In addition, to generate matrix $\mathcal{M}_b$, the customized \texttt{CNN} network employs 5 convolutional layers with 3 $\times$ 3 convolutional kernels.
To tokenize $\text{I}_{G}$, we use a \texttt{CNN} stem, which is just a convolutional layer with a large-scale convolution kernel (16 $\times$ 16).

\section{Experiments}
We introduce a set of experiments to evaluate and analyze the effectiveness of our proposed method.
In addition, we create a large dataset (\textbf{Multi-exposure Document Datasets}) with UHD resolution (resolution greater than 4K).
%
%

\noindent \textbf{Dataset.}
We evaluate the algorithm on the \textit{SICE}~\cite{cai2018learning} and \textit{NTIRE workshop22 Multi-frame HDR}~\cite{whdr} datasets.
Each of the two datasets with 30\% of the samples serves as the test set.
Each scenario in the dataset includes three LDR images with various exposure levels (short/medium/long exposure).
Moreover, since document recognition and detection is a very important application, a multi-exposure document dataset (MED) with UHD resolution is created for this purpose.
Each document contains a pair of time-aligned short-exposure and long-exposure images and their corresponding ground truth.
The underexposed and overexposed images of the document image can be formulated as follows according to the method proposed by Lv et al.~\cite{LvLL21} :
\begin{equation}
	I_{out}^{(i)} = \beta \times (\alpha \times I_{in}^{(i)})^\gamma , i \in {\{R,G,B\}}, 
\end{equation}
\noindent
where $\alpha$ and $\beta$ are linear transformations, the $\text{X}^\gamma$ ( denotes all pixels in an image) means the gamma transformation.
The three parameters are sampled from the uniform distribution $\mathbb{U}$:
$\alpha \sim \mathbb{U}(0.9,1),\beta \sim \mathbb{U}(0.5,1),\gamma \sim \mathbb{U}(1.5,5)$.
However, this customized manner of building images is difficult to meet complex environmental transformations, such as blur, and noise.
To tackle this problem, we attempt to employ CycleGAN~\cite{cycle} to enforce the noise of the environment on the synthesized dataset.
The unpaired image dataset uses \textit{SICE} (short/long images).
We utilize an AdamW optimizer with an initial learning rate of 0.001 and train the CycleGAN for 100 epochs, delaying the learning rate by 0.1 after 50 and 75 epochs.
Since CycleGAN may over-modify some images, such as generating dark corners, ghost textures, and other problems, we use adobe photoshop to repair them.
Note that we do not employ professional devices for multi-exposure image synthesis for two reasons: i) existing devices cannot meet resolutions above 6K;
ii) UHD images are difficult to align in the time dimension.
Our document dataset includes contracts, papers, invoices, and other types, with a total of 1000 sets of images.
\begin{table*}[!htb]
	\begin{center}
		\caption{We compare SSIM and PSNR indicators on three datasets respectively. In addition, we statistics the number of parameters of the models. Our method contains only 0.52M parameters.}
		\vspace{-1mm}
		\label{table:headings}
		\scriptsize
		\resizebox{\linewidth}{!}{
			\begin{tabular}{ccccccccccc}
				\toprule	
				Dataset & Metrics   &Short&Medium &Long& HALDER(-) & AGAL(24M) & AHDR(1.44M)  & U2fusion(-) &ADNet(2.80M) &\textbf{Ours}(0.52M)  \\ \midrule
				& PSNR &6.68&11.52&14.7&19.8& 18.84  & 18.2& 16.9  &21.23&\textbf{23.8} \\
				\multirow{2}{*}[2.2ex]{SICE}& SSIM &0.13&0.59&0.74 & 0.82 & 0.83 & 0.71&0.79 &\textbf{0.82} &0.81\\  \midrule
				
				& PSNR & 13.50 &34.2& 15.09 & 35.33 & 26.57&35.11 &30.2&35.78&\textbf{36.85} \\
				\multirow{2}{*}[2.2ex]{NTIRE22}  & SSIM &0.49   & 0.83& 0.69 & 0.85 & 0.69 & 0.92&0.61& 0.93 &\textbf{0.95}\\  \midrule

				& PSNR  &4.24&-&10.28& 19.31 &18.22 &27.14 &25.31 & 26.88 &   \textbf{29.78} \\
				\multirow{2}{*}[2.2ex]{MED}  & SSIM &0.08&-& 0.42 & 0.71 & 0.69 &0.92 &0.85& 0.94&\textbf{0.97}\\ 
				\bottomrule
		\end{tabular} }
		\vspace{-2mm}
	\end{center}
\end{table*}

\begin{figure*}[!h]\scriptsize
	\tabcolsep 1pt
	\begin{tabular}{@{}ccccc@{}}
		\includegraphics[width = 0.2\textwidth]{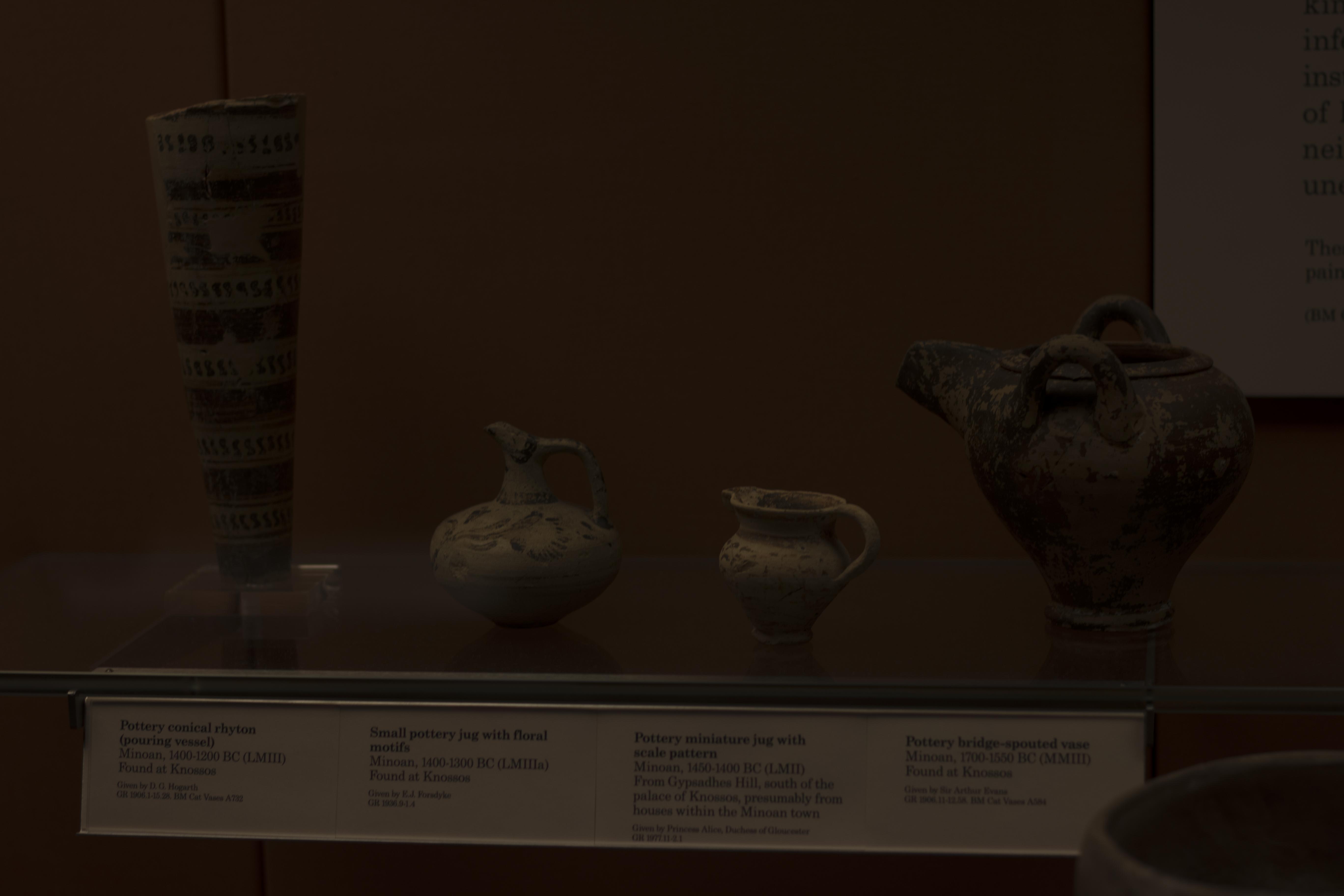}              &
		\includegraphics[width = 0.2\textwidth]{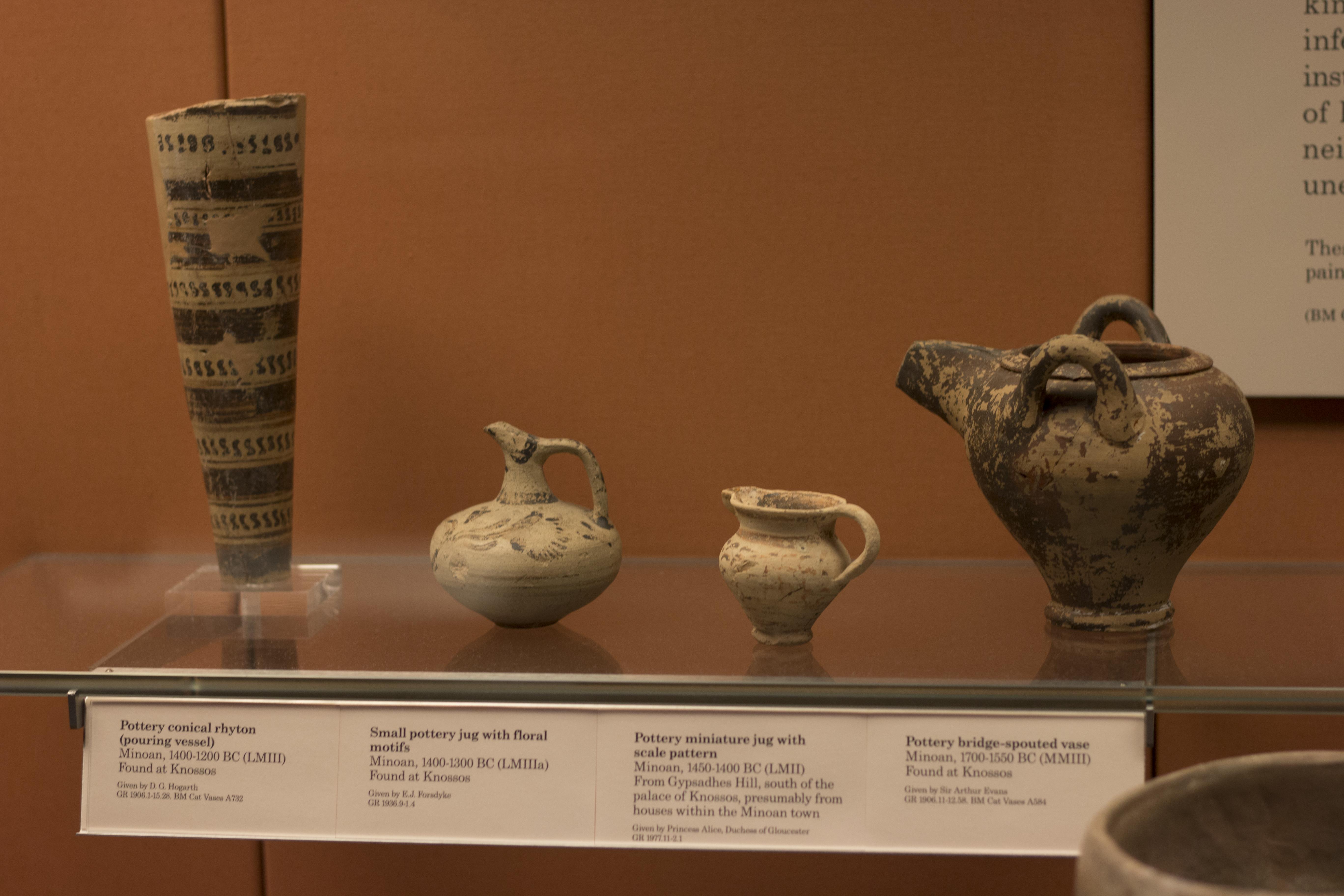}               &
		\includegraphics[width = 0.2\textwidth]{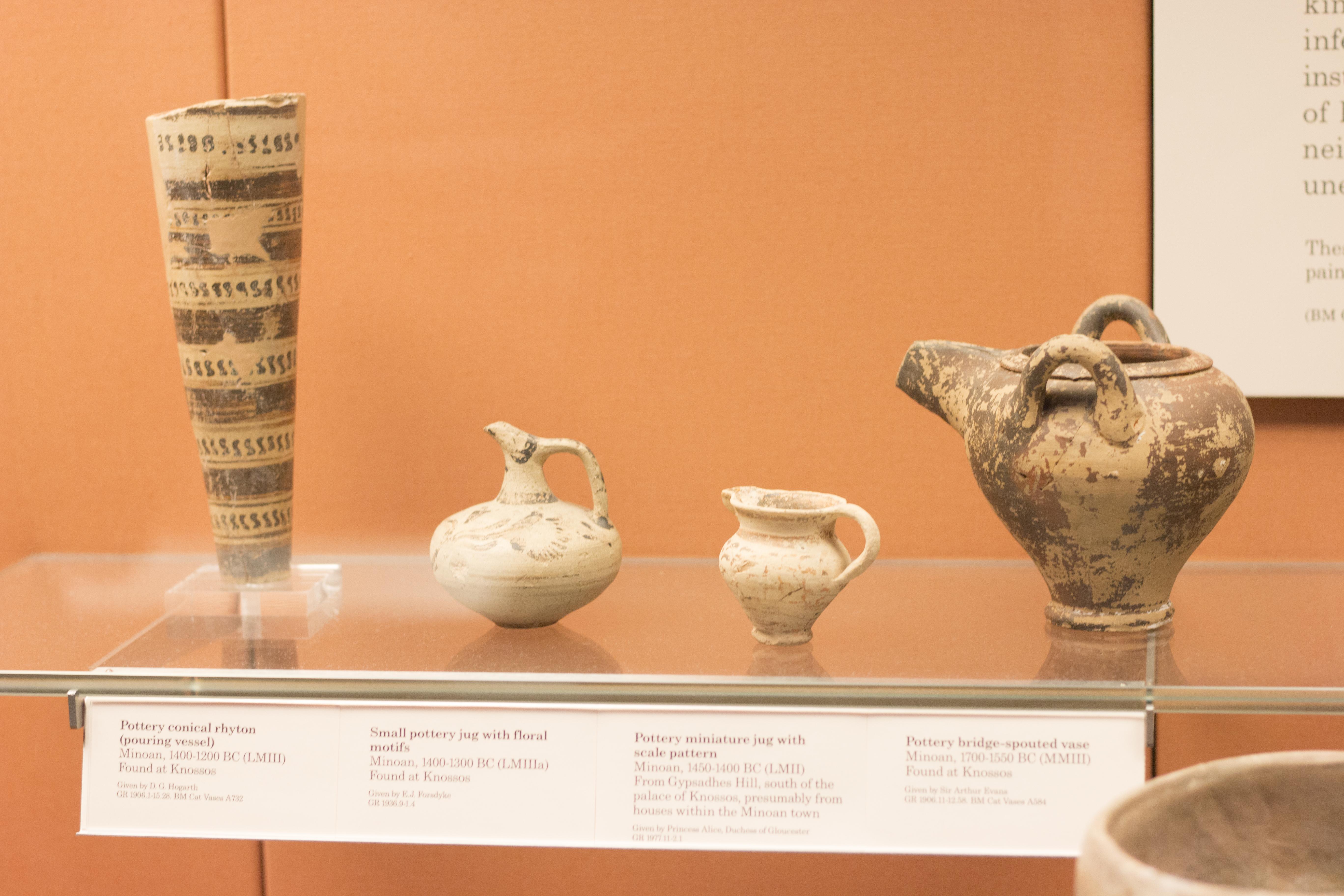}                &
		\includegraphics[width = 0.2\textwidth]{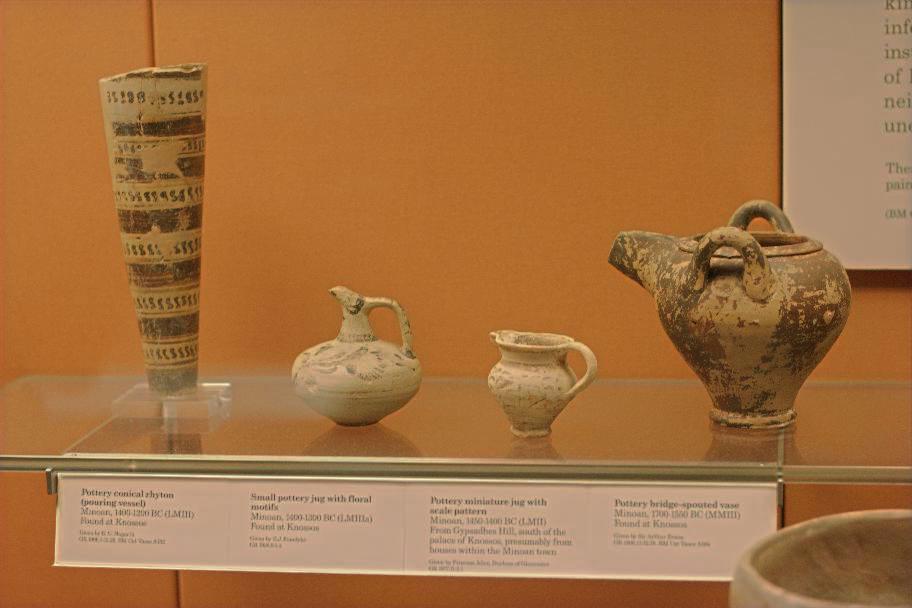}                 &
		\includegraphics[width = 0.2\textwidth]{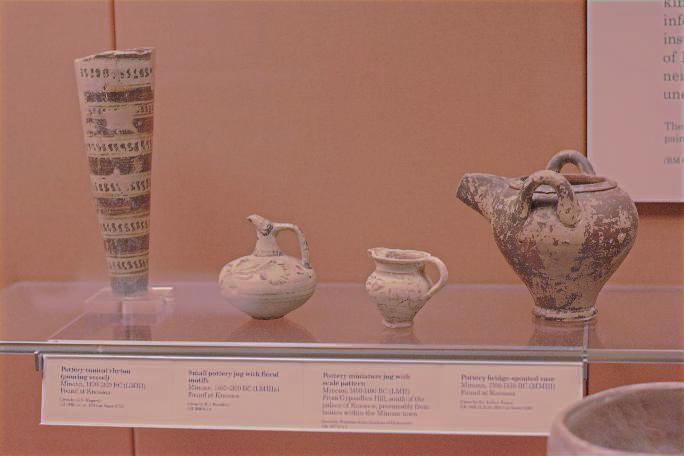}                       \\
		
		\includegraphics[width = 0.2\textwidth]{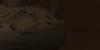}              &
		\includegraphics[width = 0.2\textwidth]{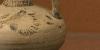}               &
		\includegraphics[width = 0.2\textwidth]{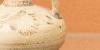}                &
		\includegraphics[width = 0.2\textwidth]{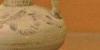}                 &
		\includegraphics[width = 0.2\textwidth]{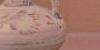}                       \\
		
		Short &
		Medium~& 
		Long~& 
		HALDER~& 
		AGAL~\\
		
		\includegraphics[width = 0.2\textwidth]{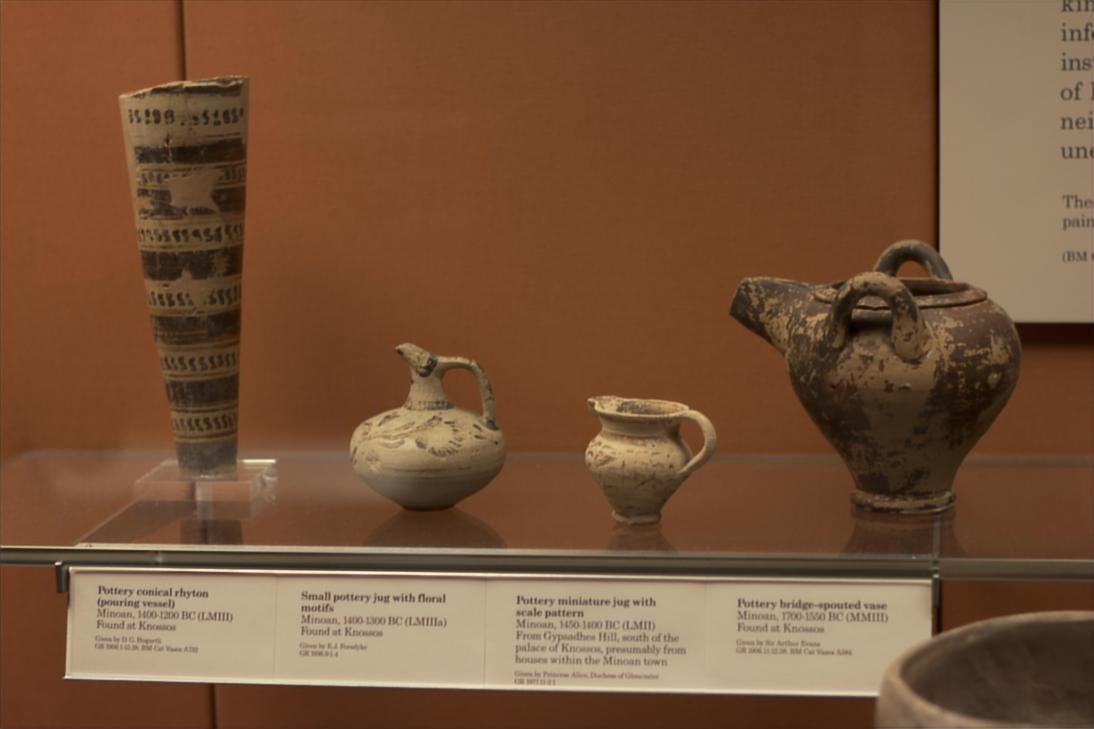}             & 
		\includegraphics[width = 0.2\textwidth]{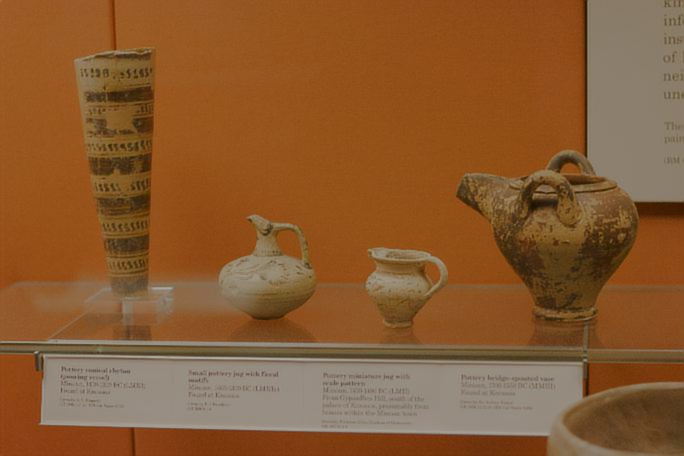}       &
		\includegraphics[width = 0.2\textwidth]{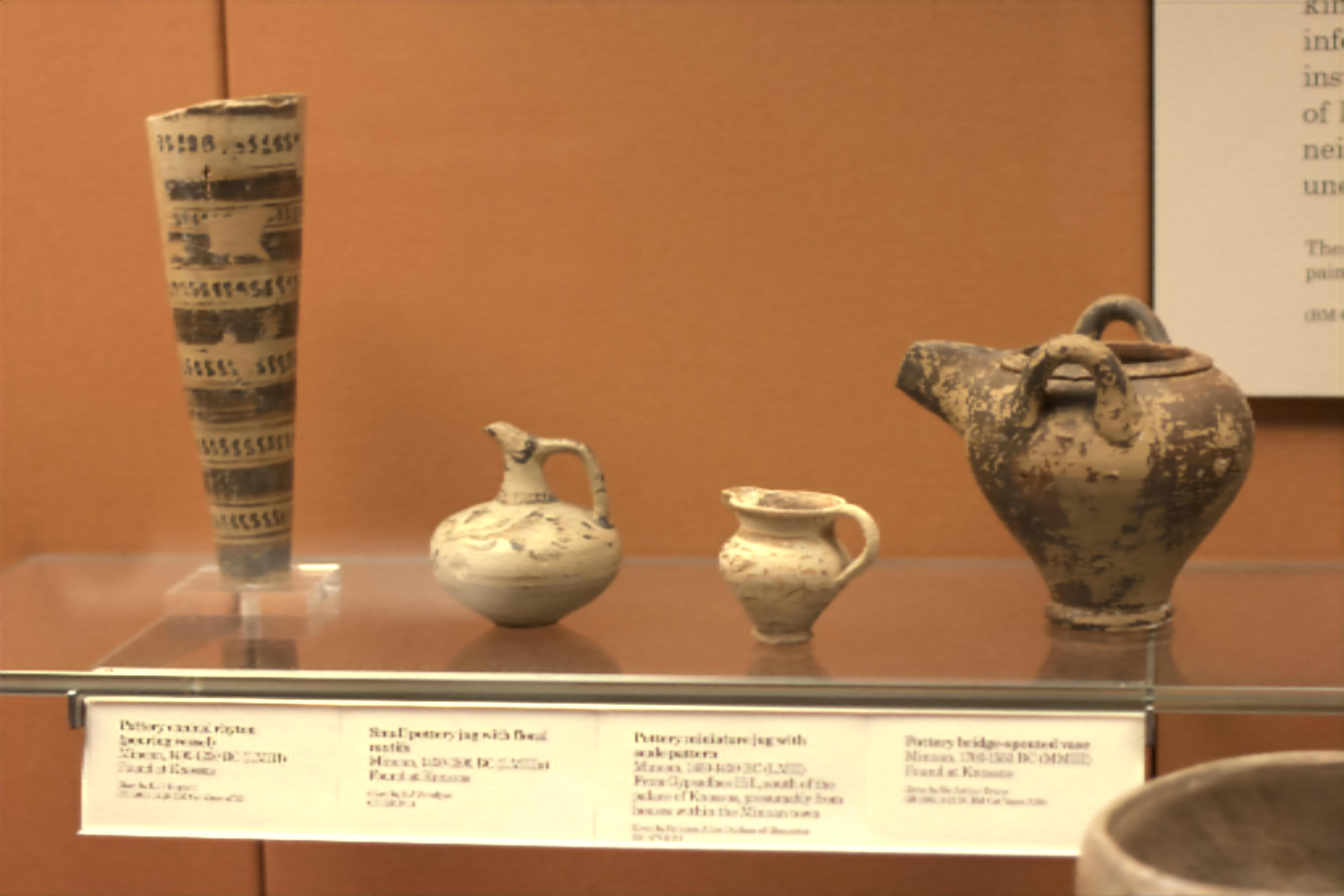}              &
		\includegraphics[width = 0.2\textwidth]{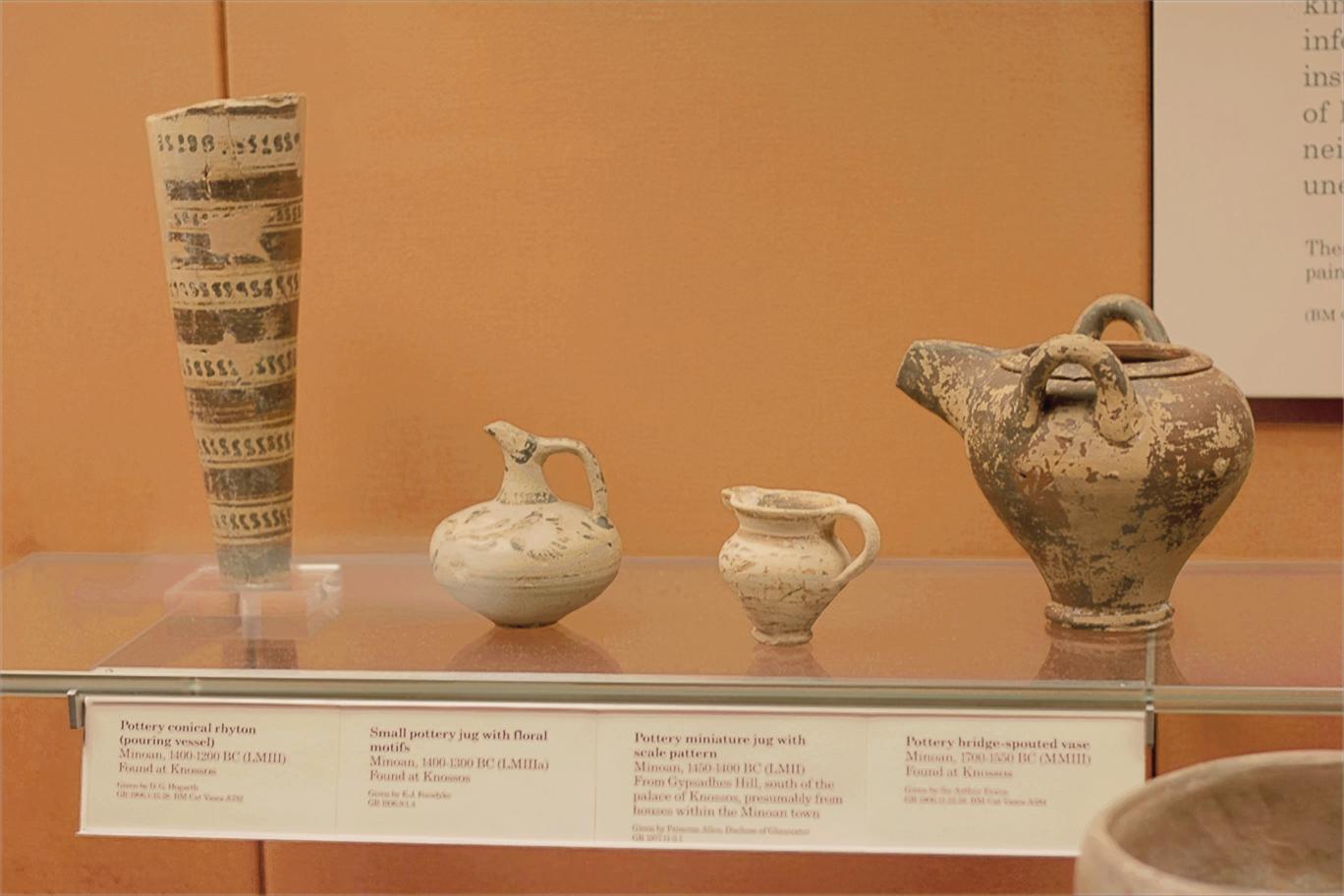}                 & 
		\includegraphics[width = 0.2\textwidth]{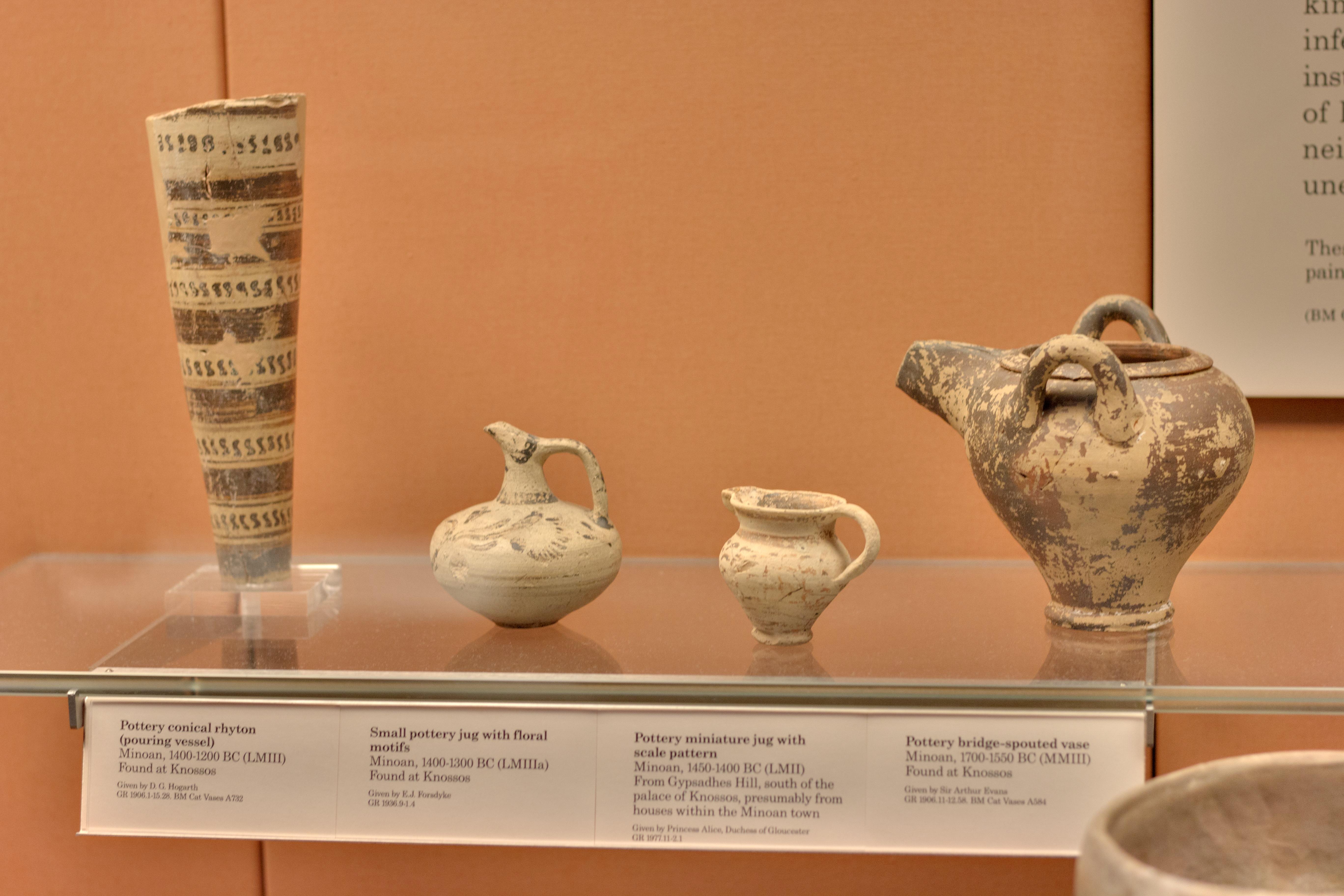}                     \\   
		
		\includegraphics[width = 0.2\textwidth]{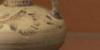}             & 
		\includegraphics[width = 0.2\textwidth]{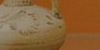}       &
		\includegraphics[width = 0.2\textwidth]{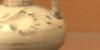}              &
		\includegraphics[width = 0.2\textwidth]{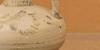}                 & 
		\includegraphics[width = 0.2\textwidth]{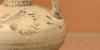}                     \\

		AHDR~& 
		U2fusion~& 
		ADNet~ & 
		\textbf{Ours}&
		GT
		\\

		\includegraphics[width = 0.2\textwidth]{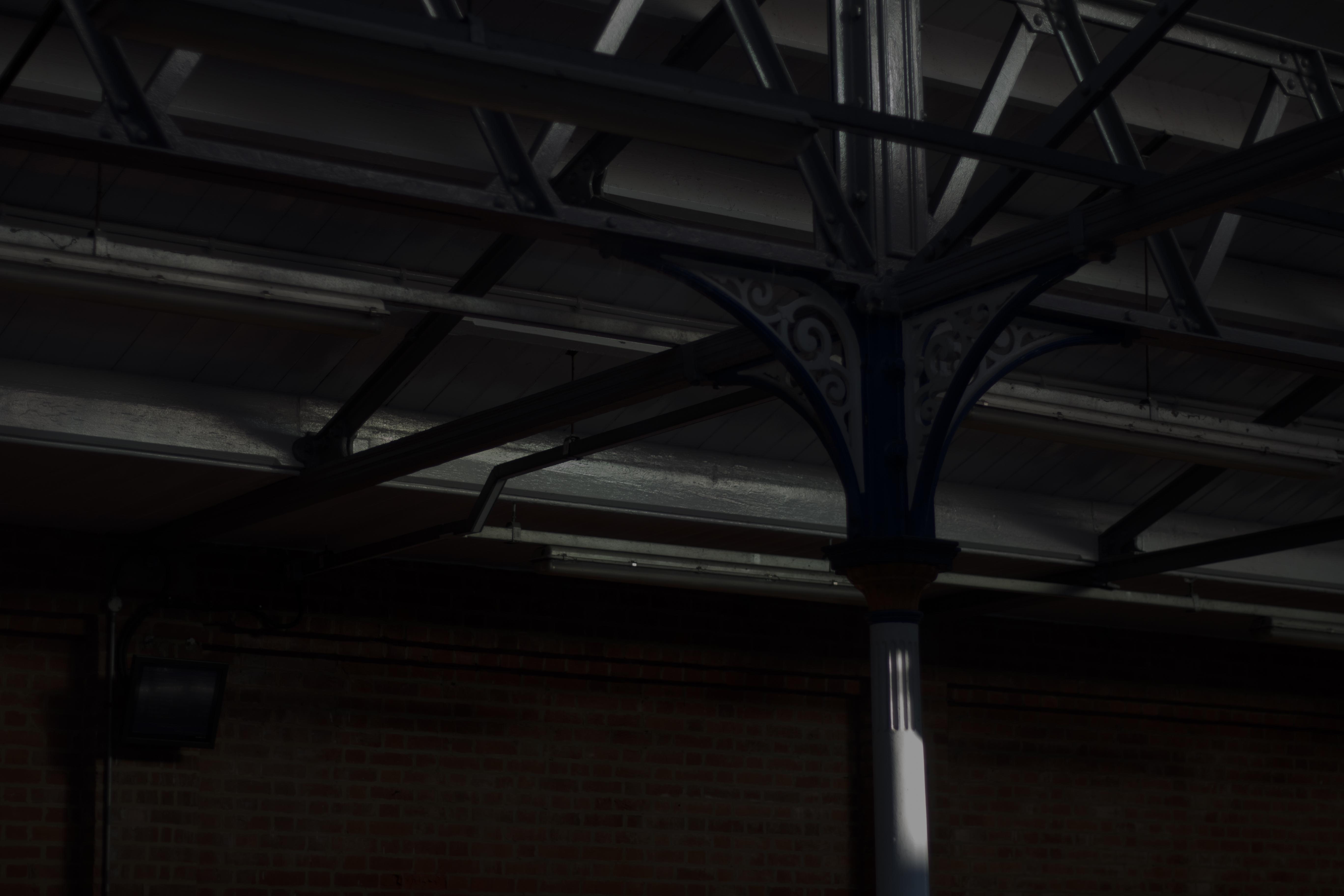}                 &
		\includegraphics[width = 0.2\textwidth]{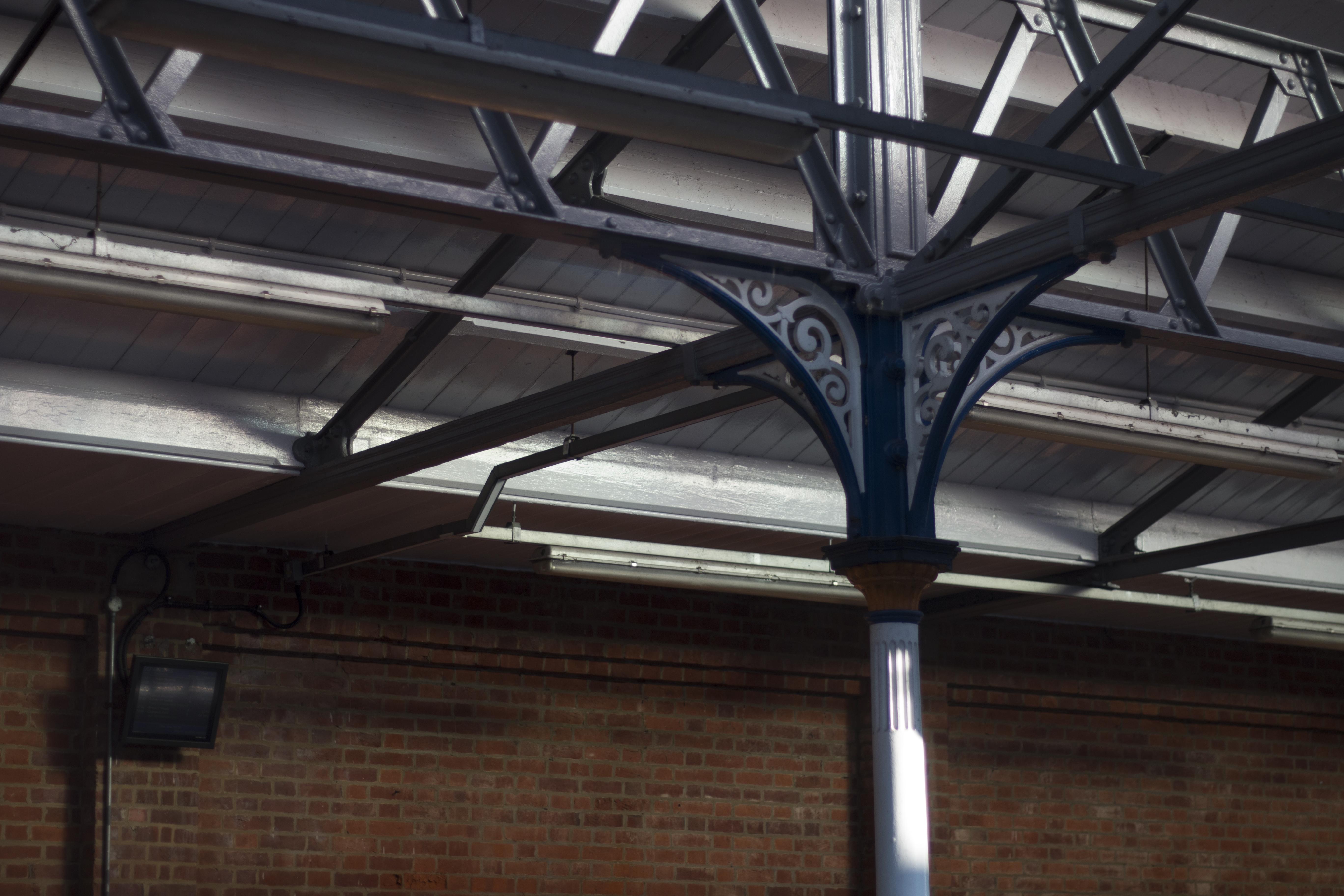}               &
		\includegraphics[width = 0.2\textwidth]{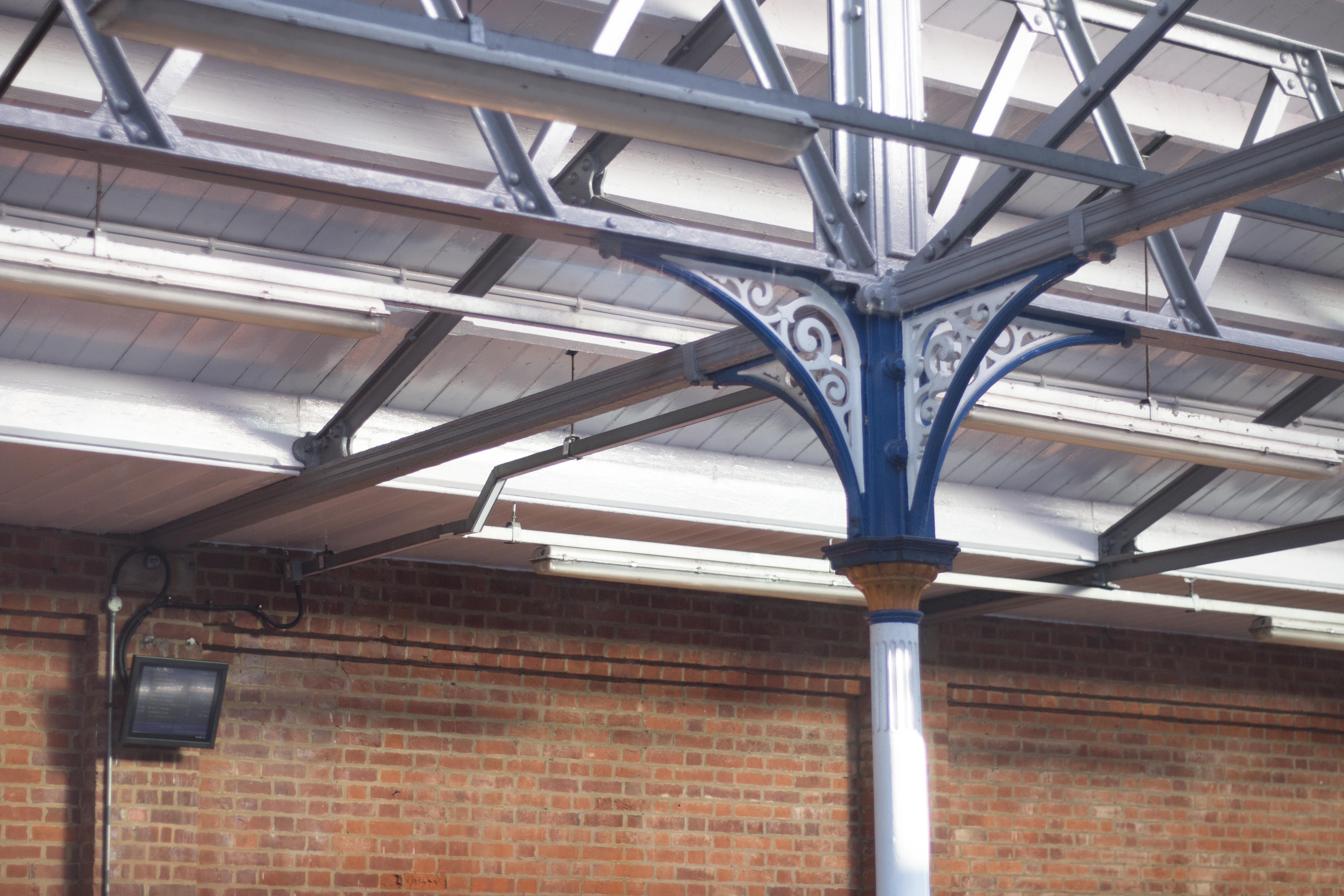}                &
		\includegraphics[width = 0.2\textwidth]{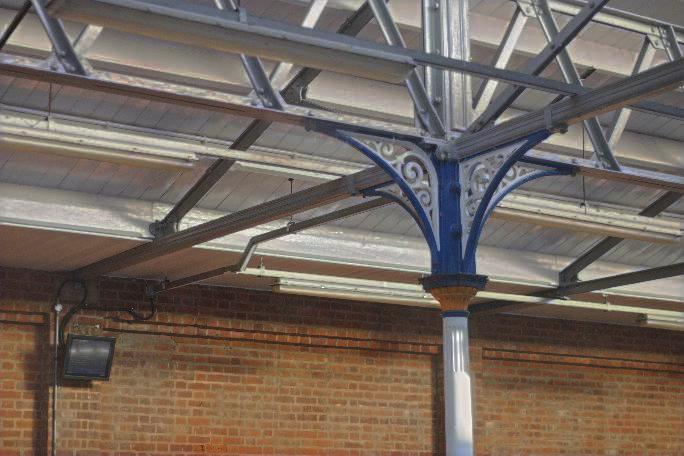}                 &
		\includegraphics[width = 0.2\textwidth]{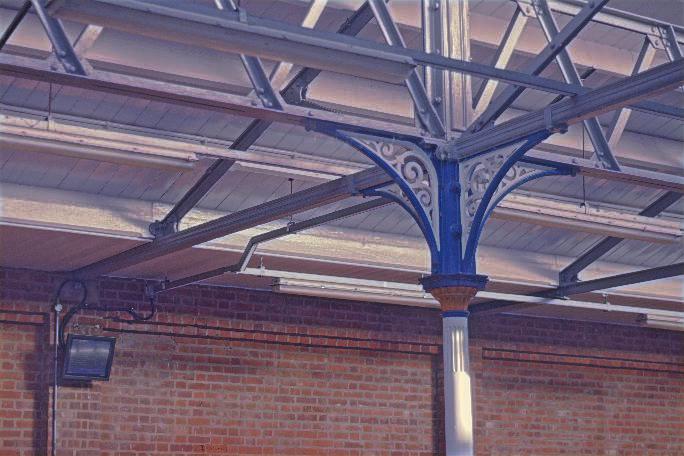}                       \\
		
		\includegraphics[width = 0.2\textwidth]{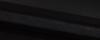}                 &
		\includegraphics[width = 0.2\textwidth]{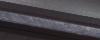}               &
		\includegraphics[width = 0.2\textwidth]{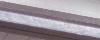}                &
		\includegraphics[width = 0.2\textwidth]{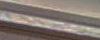}                 &
		\includegraphics[width = 0.2\textwidth]{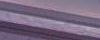}                       \\
		
		Short &
		Medium& 
		Long & 
		HALDER& 
		AGAL\\

		\includegraphics[width = 0.2\textwidth]{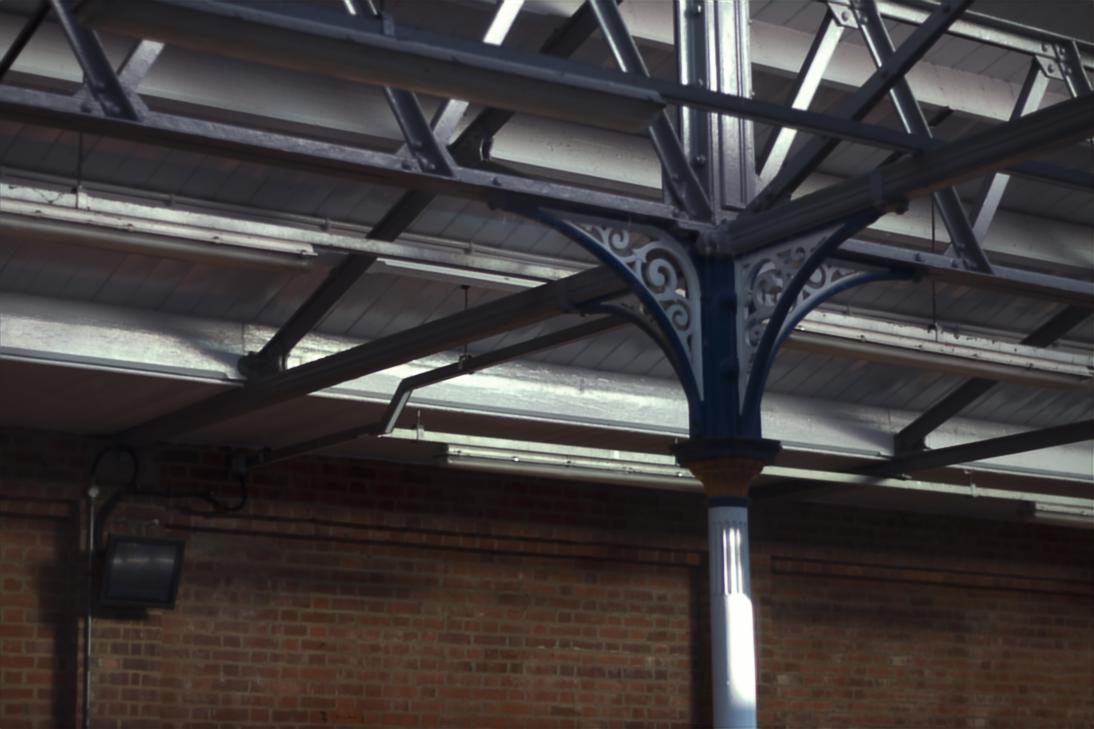}             & 
		\includegraphics[width = 0.2\textwidth]{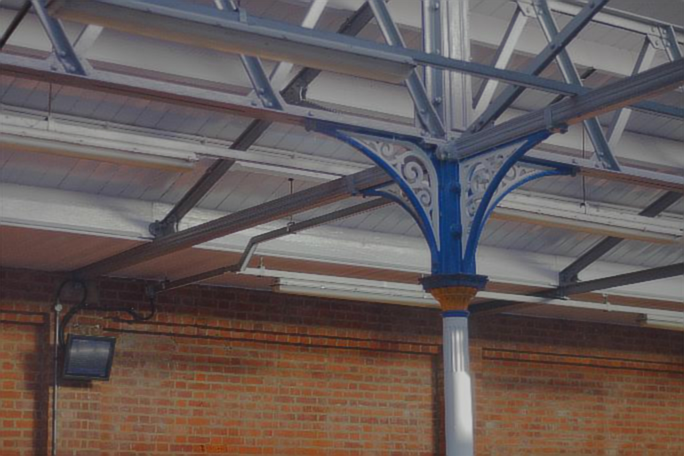}       &
		\includegraphics[width = 0.2\textwidth]{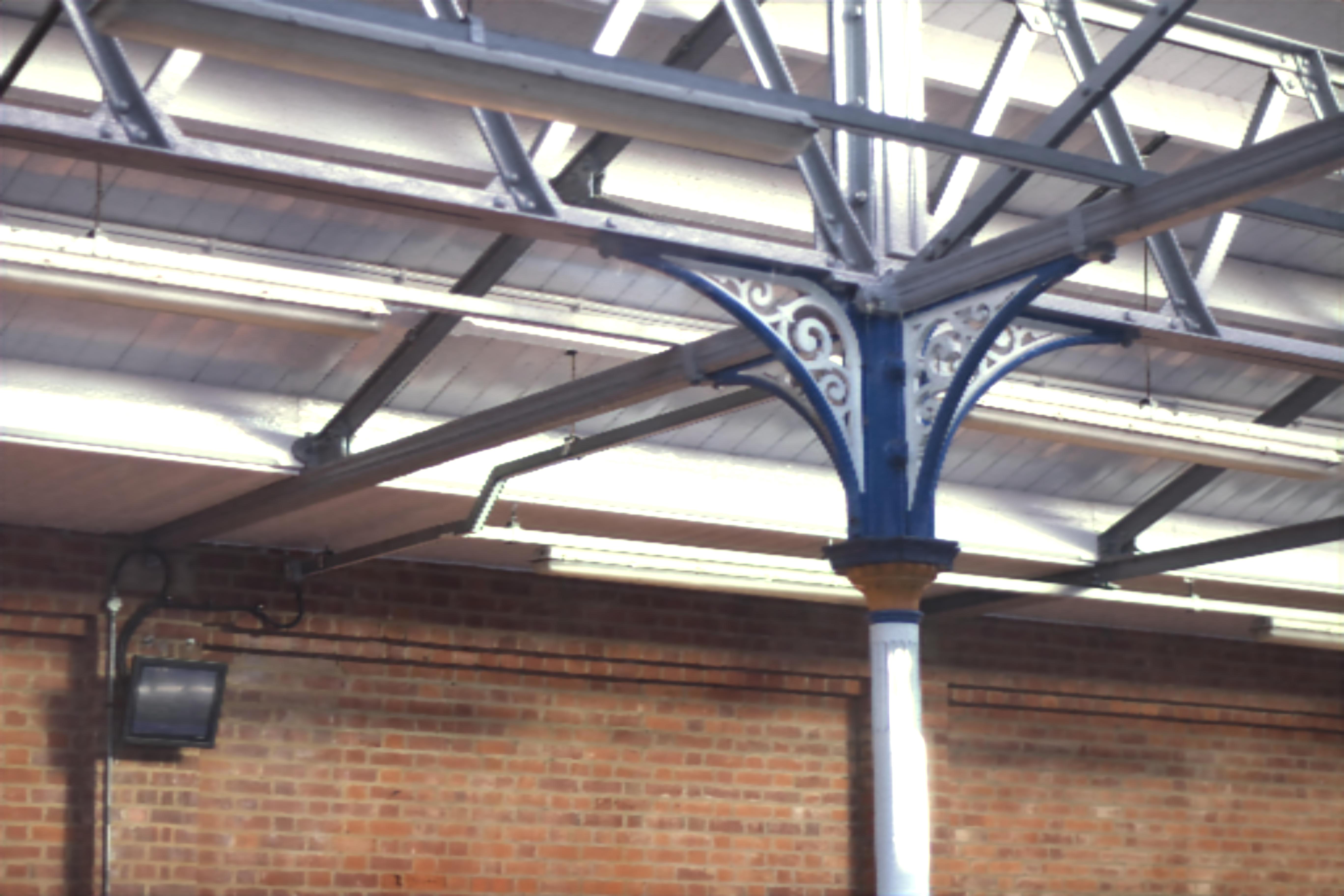}                & 
		\includegraphics[width = 0.2\textwidth]{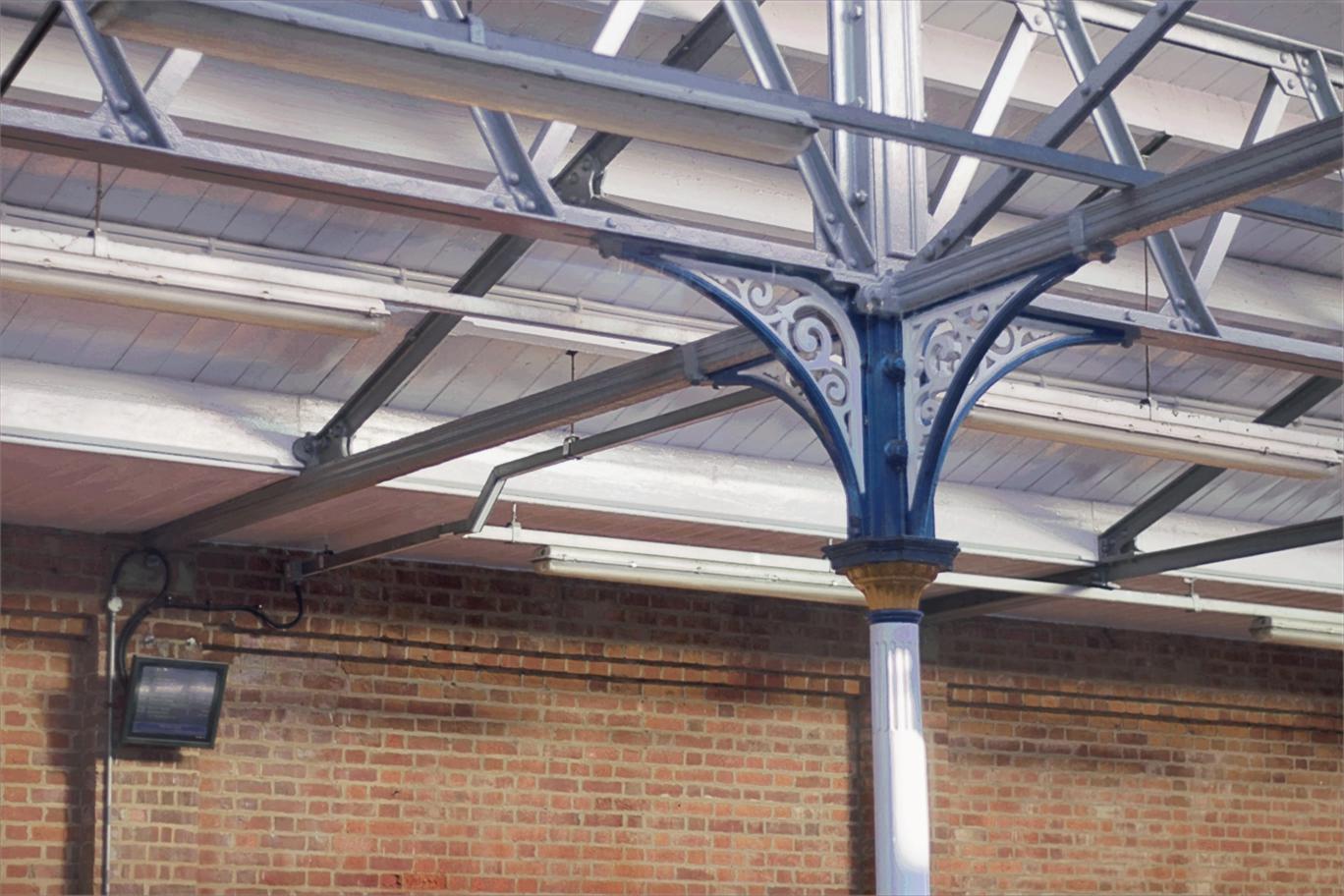}                 & 
		\includegraphics[width = 0.2\textwidth]{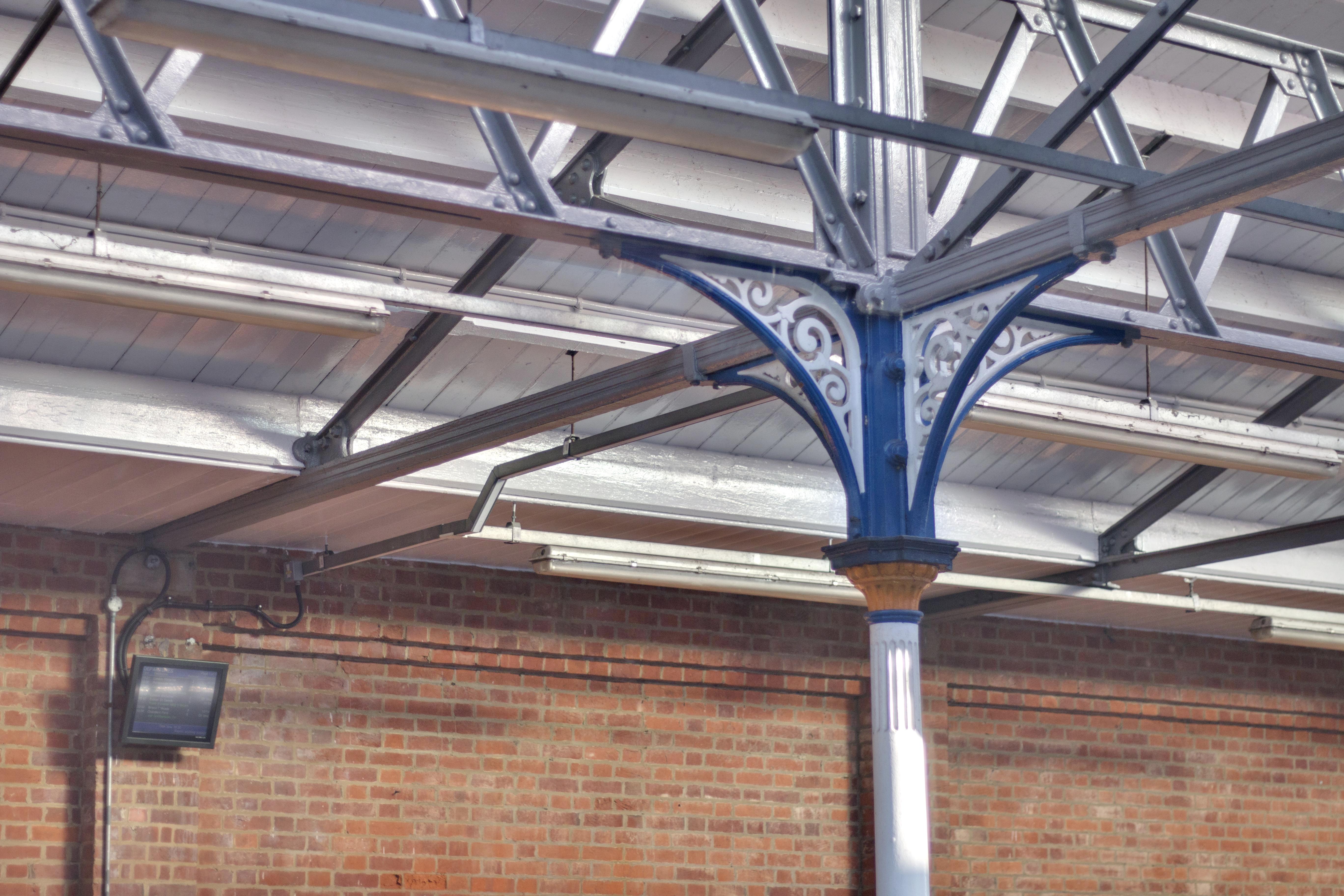}                     \\ 
		
		\includegraphics[width = 0.2\textwidth]{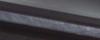}                 &
		\includegraphics[width = 0.2\textwidth]{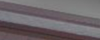}               &
		\includegraphics[width = 0.2\textwidth]{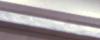}                &
		\includegraphics[width = 0.2\textwidth]{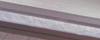}                 &
		\includegraphics[width = 0.2\textwidth]{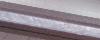}                       \\

		AHDR&
		U2fusion & 
		ADNet & 
		\textbf{Ours}&
		GT 
		\\

	\end{tabular}
	
	\caption{Our method obtains better visual quality and recovers more image details compared with other state-of-the-art methods in the SICE datasets.}	
	\label{SICE}
	\vspace{-0mm}
\end{figure*}

\begin{figure*}[t]\scriptsize
	\tabcolsep 1pt
	\begin{tabular}{@{}ccccc@{}}
		\includegraphics[width = 0.2\textwidth]{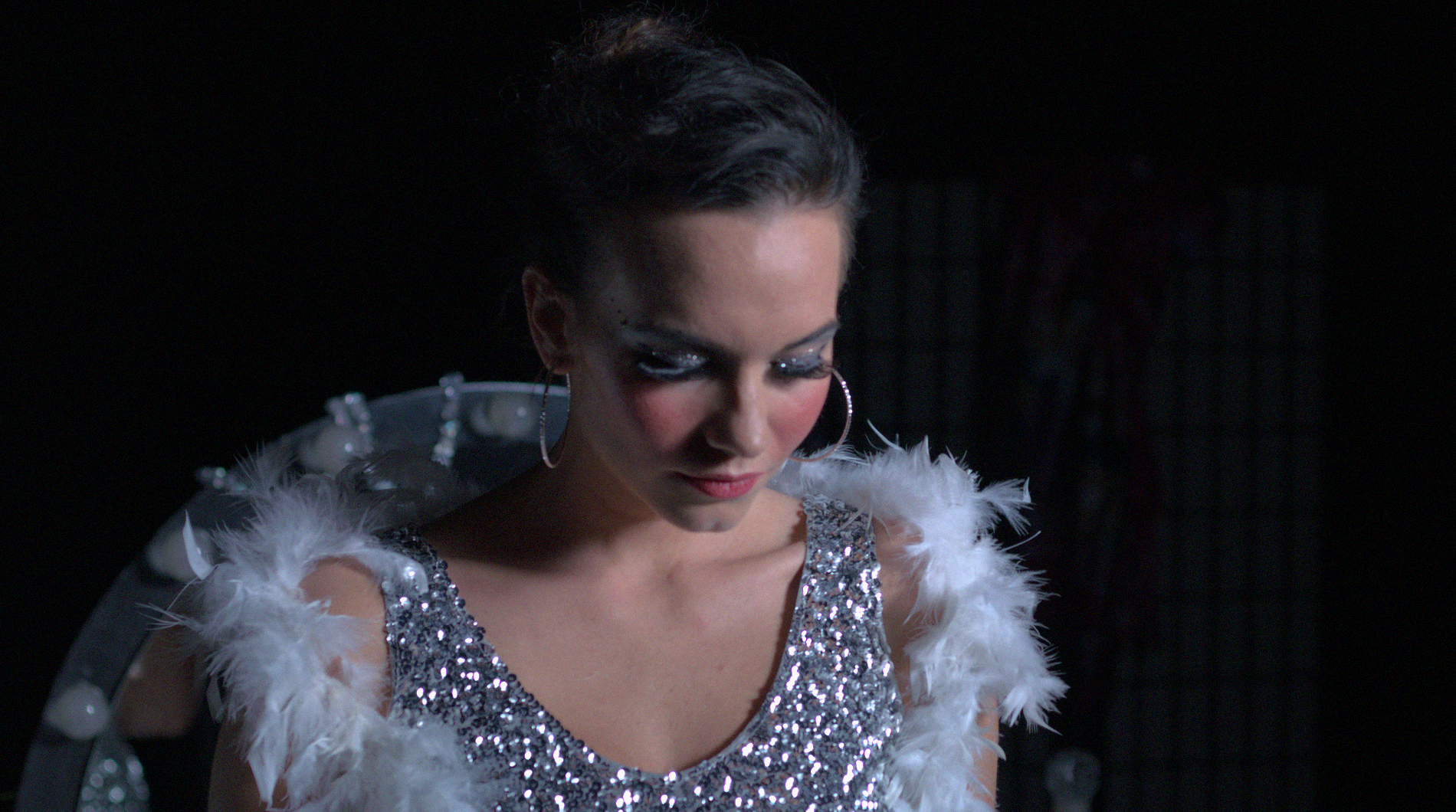}              &
		\includegraphics[width = 0.2\textwidth]{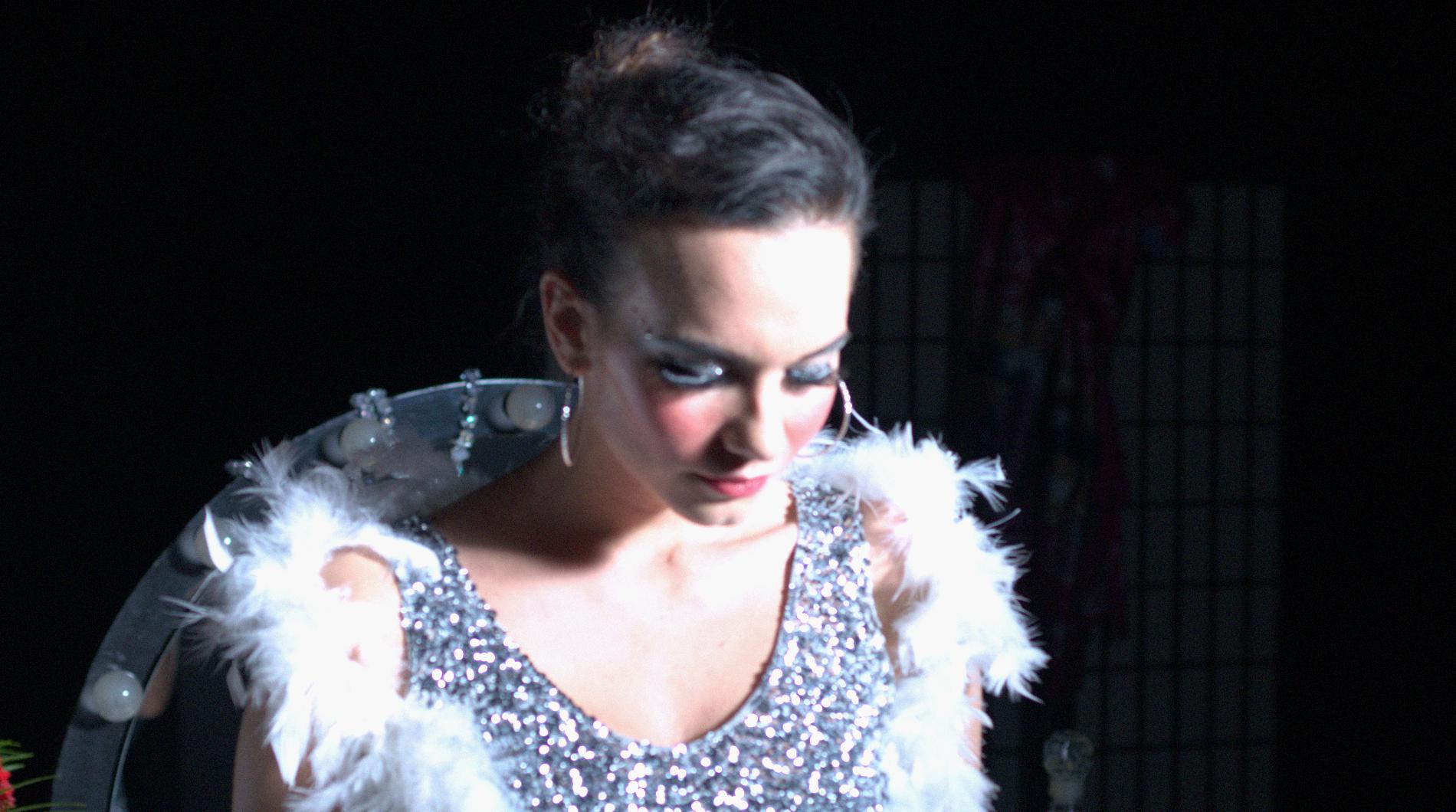}               &
		\includegraphics[width = 0.2\textwidth]{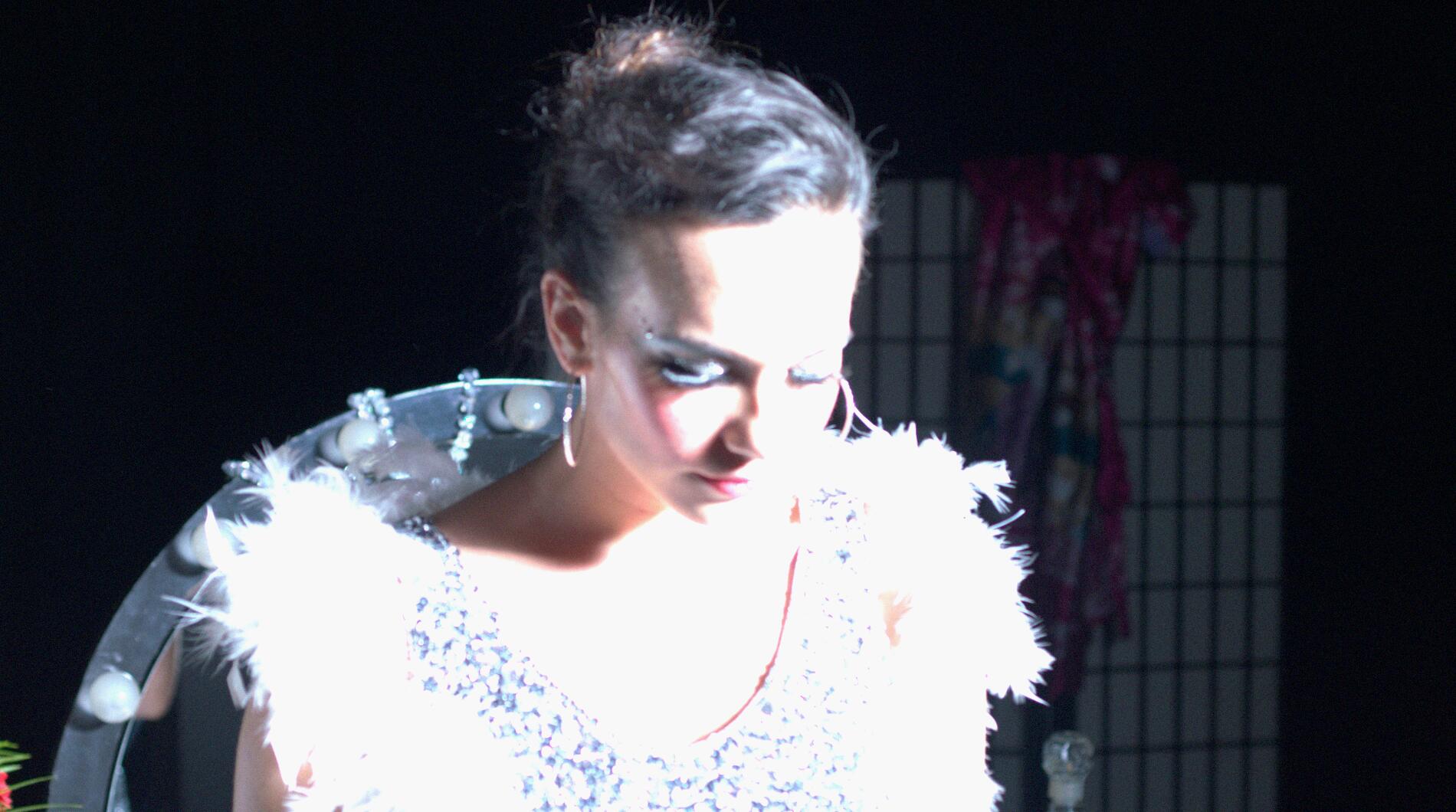}                &
		\includegraphics[width = 0.2\textwidth]{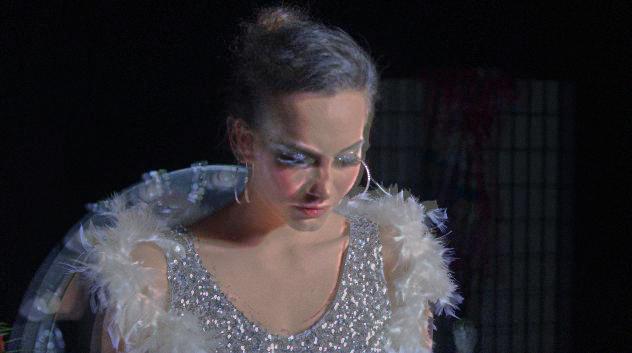}                 &
		\includegraphics[width = 0.2\textwidth]{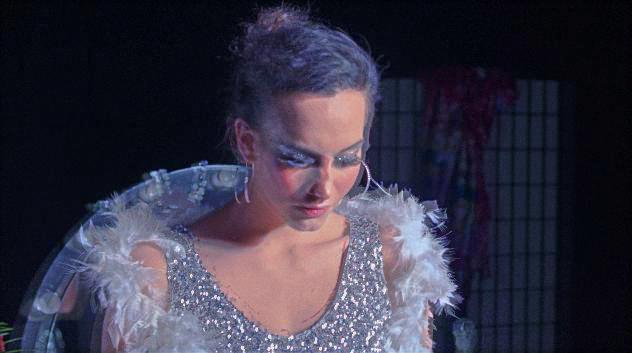}                       \\
		
		\includegraphics[width = 0.2\textwidth]{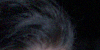}                 &
		\includegraphics[width = 0.2\textwidth]{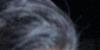}               &
		\includegraphics[width = 0.2\textwidth]{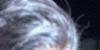}                &
		\includegraphics[width = 0.2\textwidth]{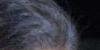}                 &
		\includegraphics[width = 0.2\textwidth]{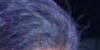}                       \\
		short &
		medium& 
		long& 
		HALDER& 
		AGAL\\
		
		\includegraphics[width = 0.2\textwidth]{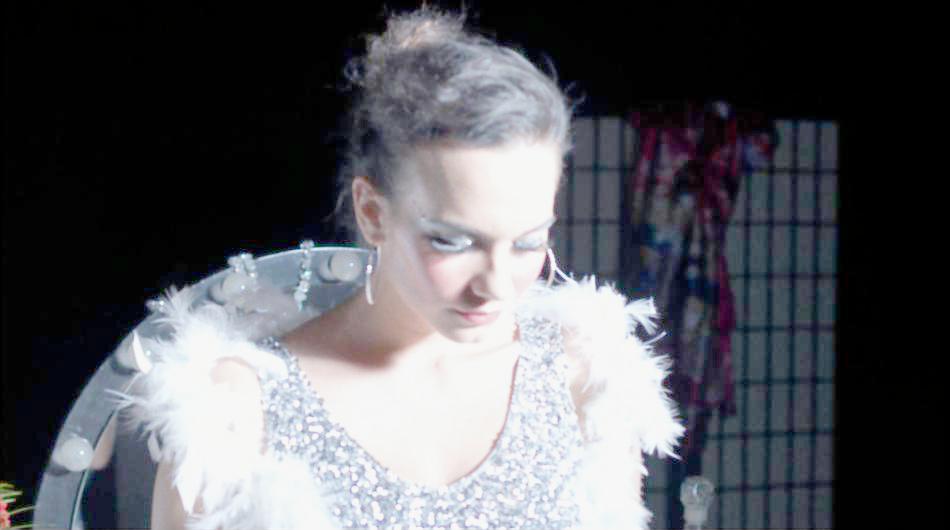}             & 
		\includegraphics[width = 0.2\textwidth]{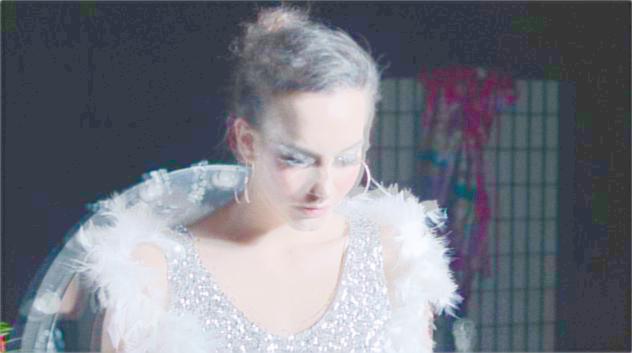}       &
		\includegraphics[width = 0.2\textwidth]{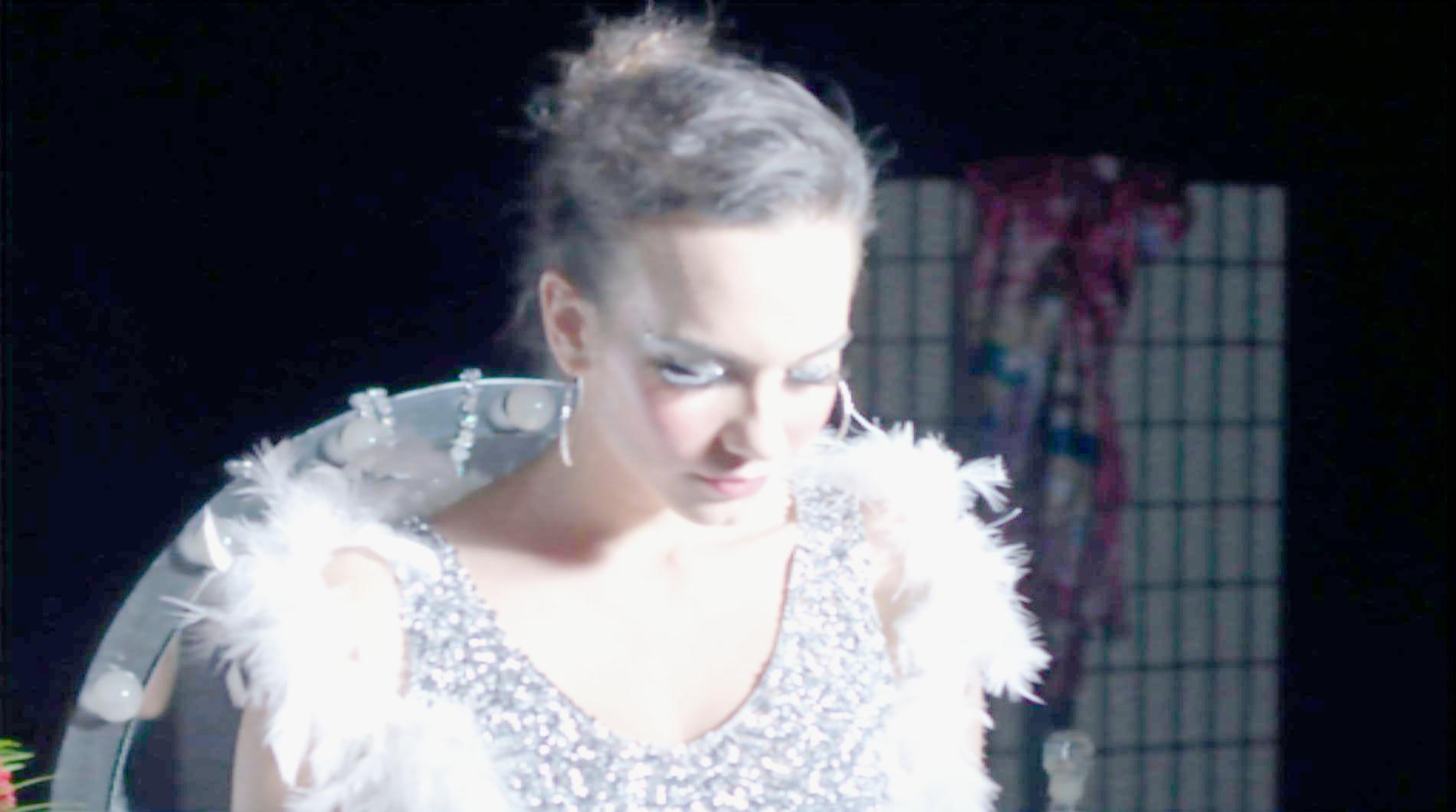}              &
		\includegraphics[width = 0.2\textwidth]{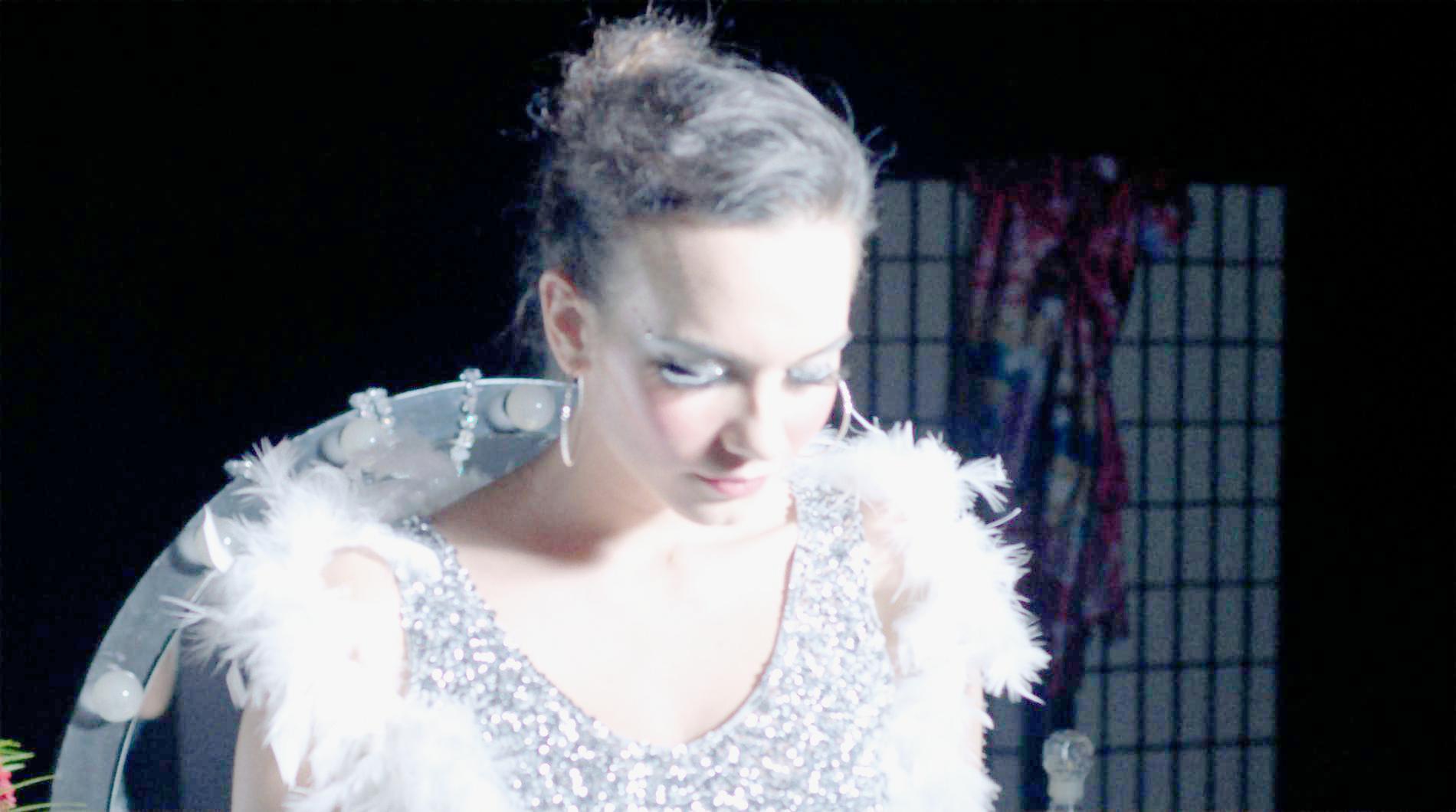}                 & 
		\includegraphics[width = 0.2\textwidth]{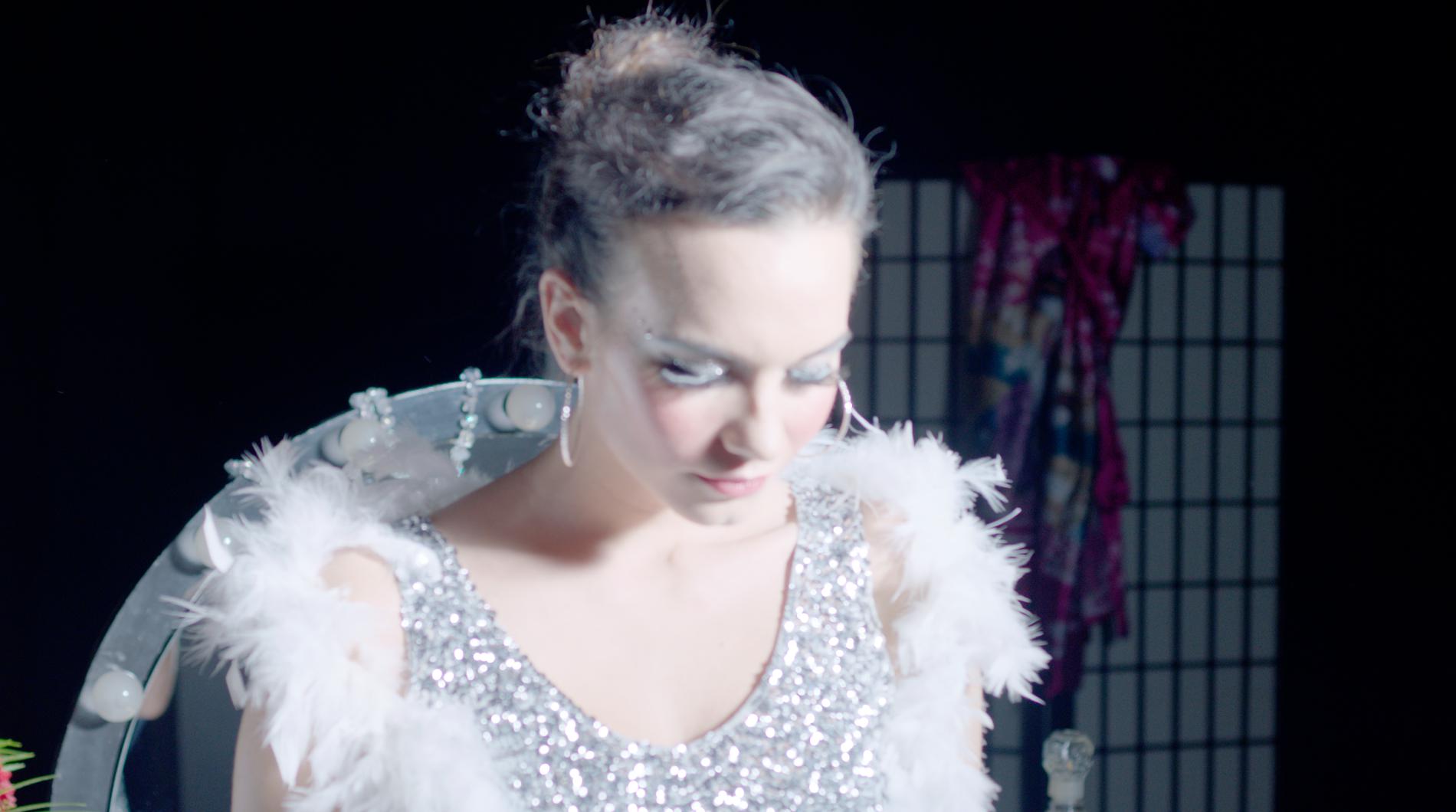}                     \\   
		
		\includegraphics[width = 0.2\textwidth]{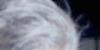}                 &
		\includegraphics[width = 0.2\textwidth]{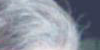}               &
		\includegraphics[width = 0.2\textwidth]{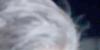}                &
		\includegraphics[width = 0.2\textwidth]{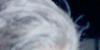}                 &
		\includegraphics[width = 0.2\textwidth]{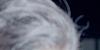}                       \\
		AHDR & 
		U2fusion& 
		ADNet & 
		\textbf{Ours}&
		GT
		\\

		\includegraphics[width = 0.2\textwidth]{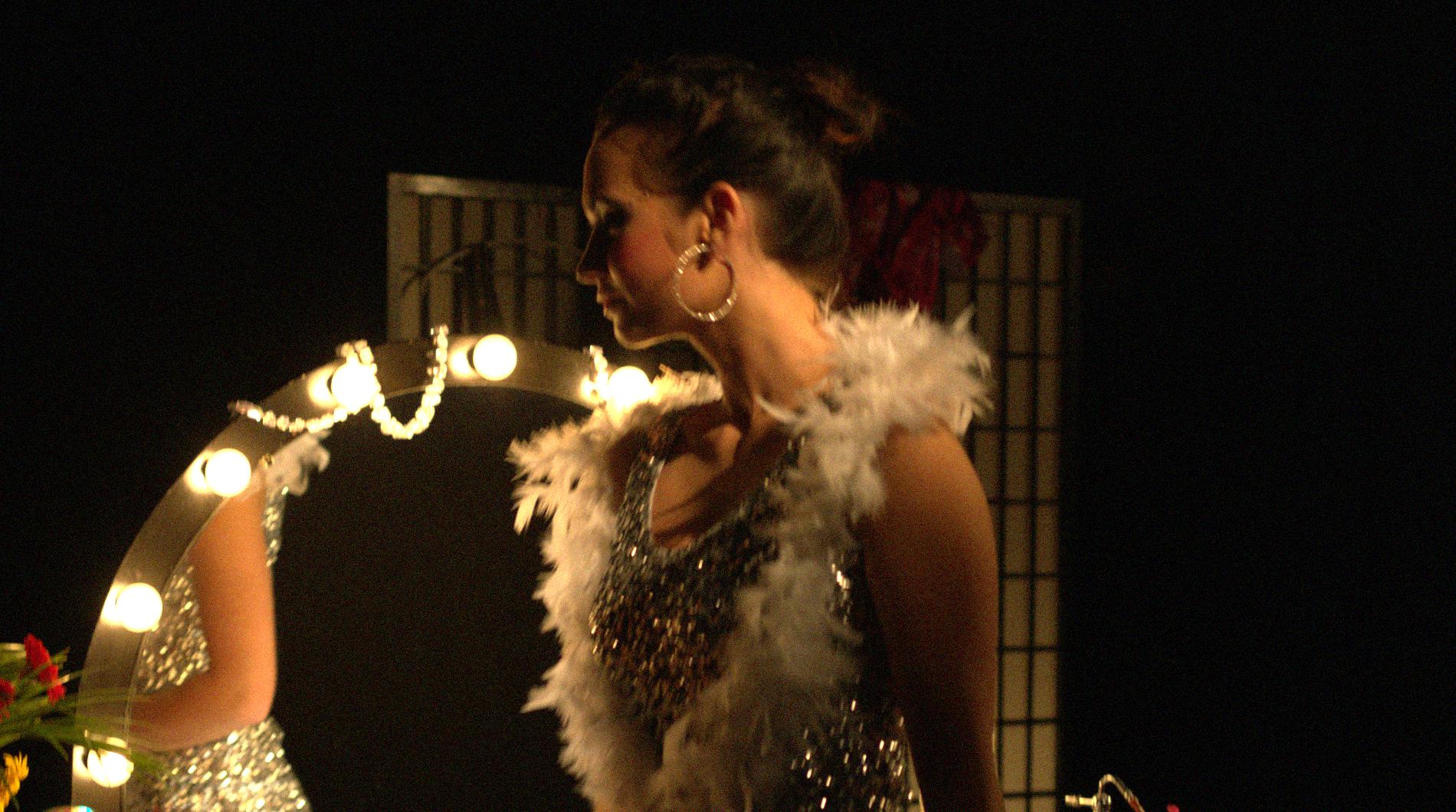}                 &
		\includegraphics[width = 0.2\textwidth]{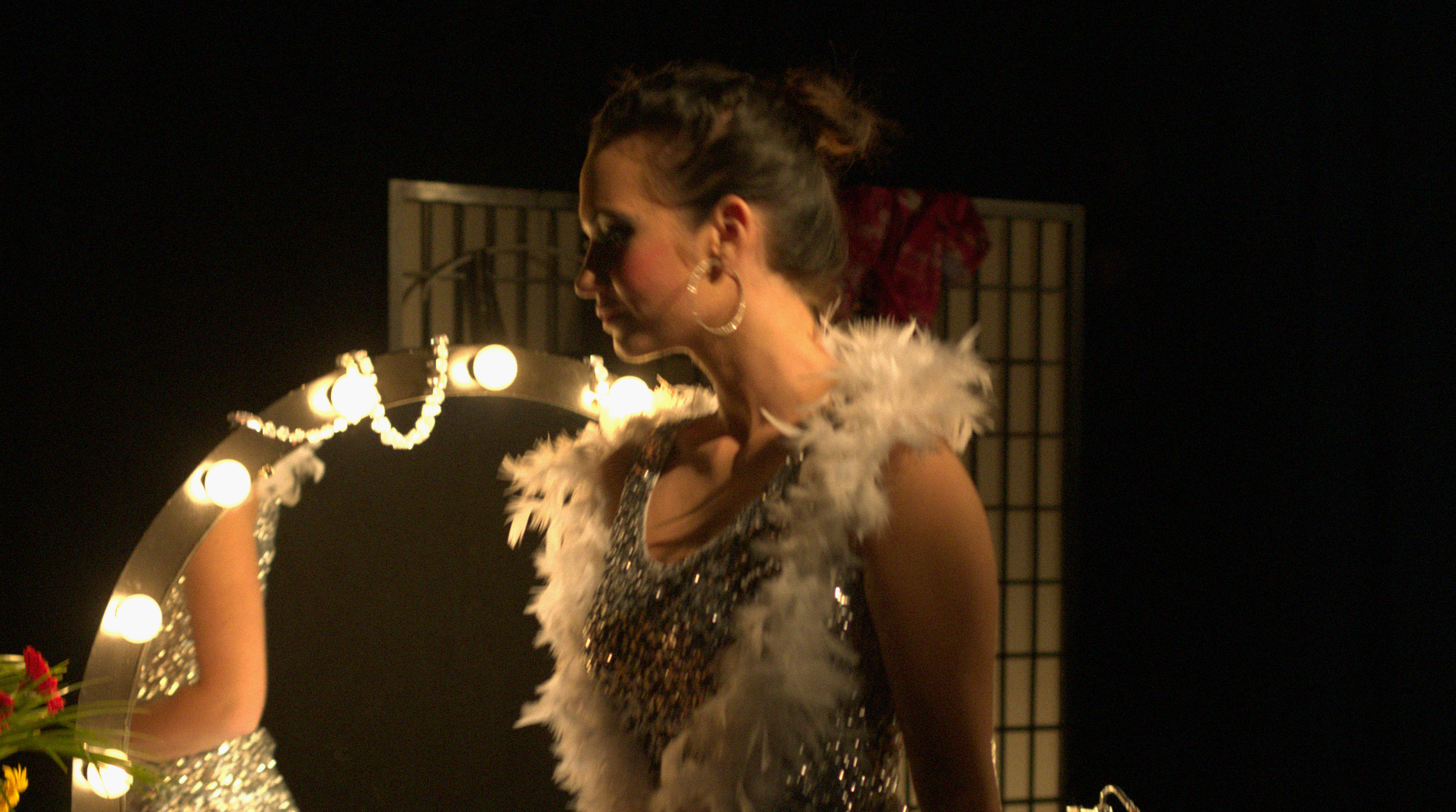}               &
		\includegraphics[width = 0.2\textwidth]{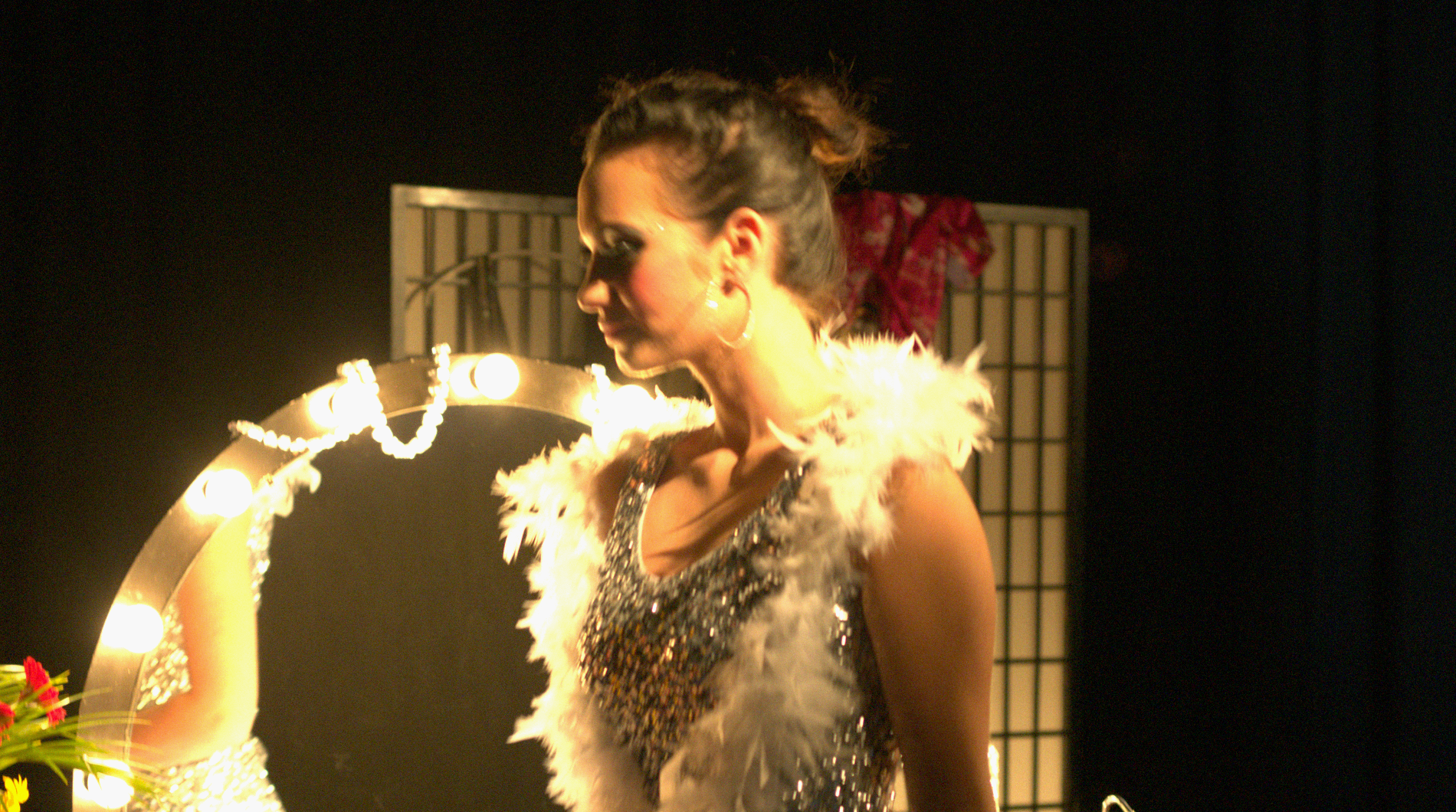}                &
		\includegraphics[width = 0.2\textwidth]{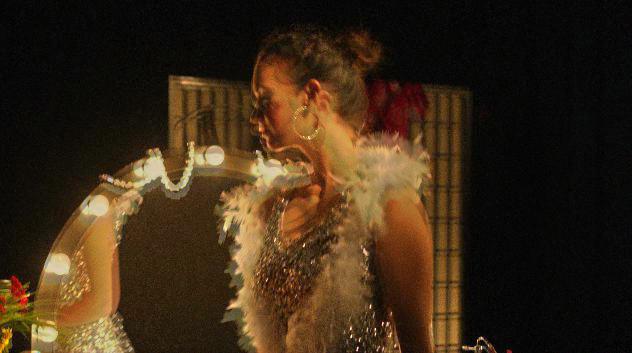}                 &
		\includegraphics[width = 0.2\textwidth]{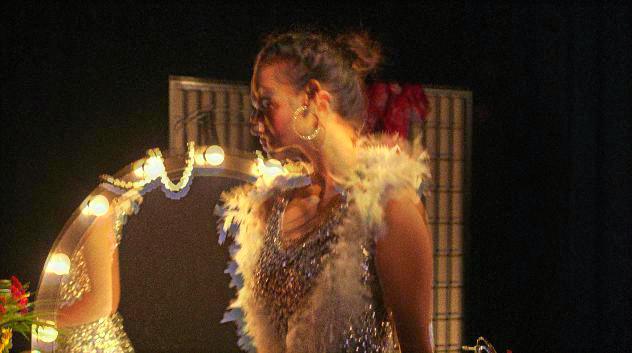}                       \\
		
		\includegraphics[width = 0.2\textwidth]{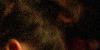}                 &
		\includegraphics[width = 0.2\textwidth]{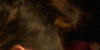}               &
		\includegraphics[width = 0.2\textwidth]{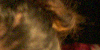}                &
		\includegraphics[width = 0.2\textwidth]{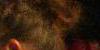}                 &
		\includegraphics[width = 0.2\textwidth]{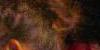}                       \\
		
		Short &
		Medium& 
		Long & 
		HALDER& 
		AGAL\\

		\includegraphics[width = 0.2\textwidth]{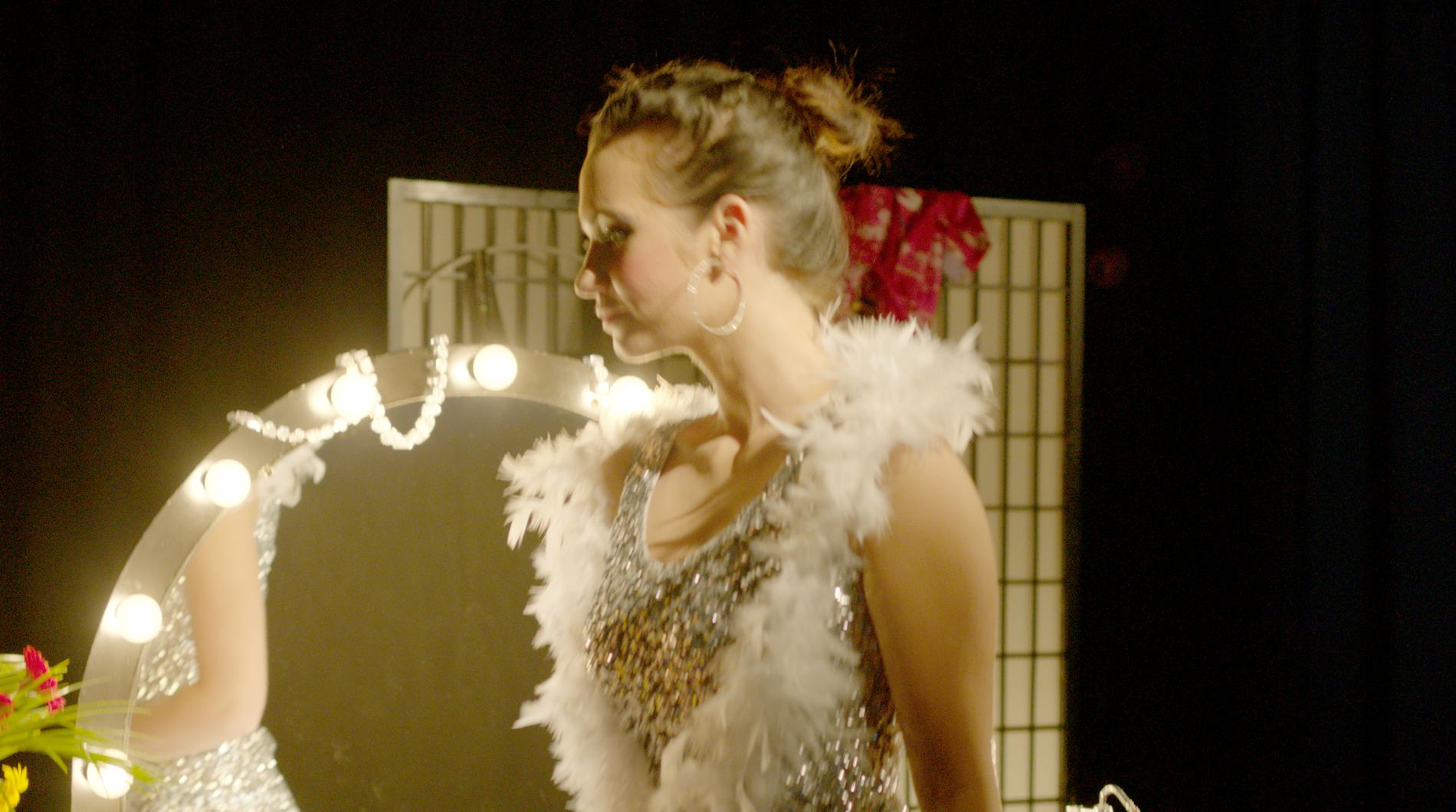}             & 
		\includegraphics[width = 0.2\textwidth]{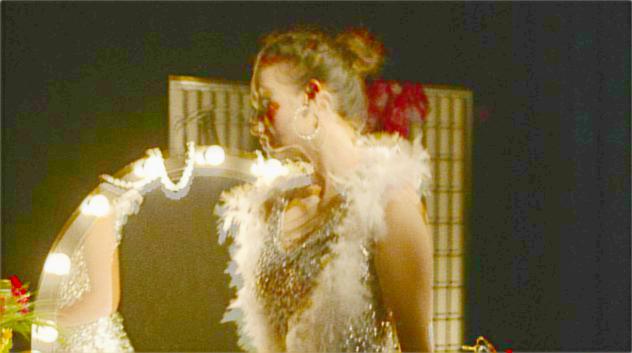}       &
		\includegraphics[width = 0.2\textwidth]{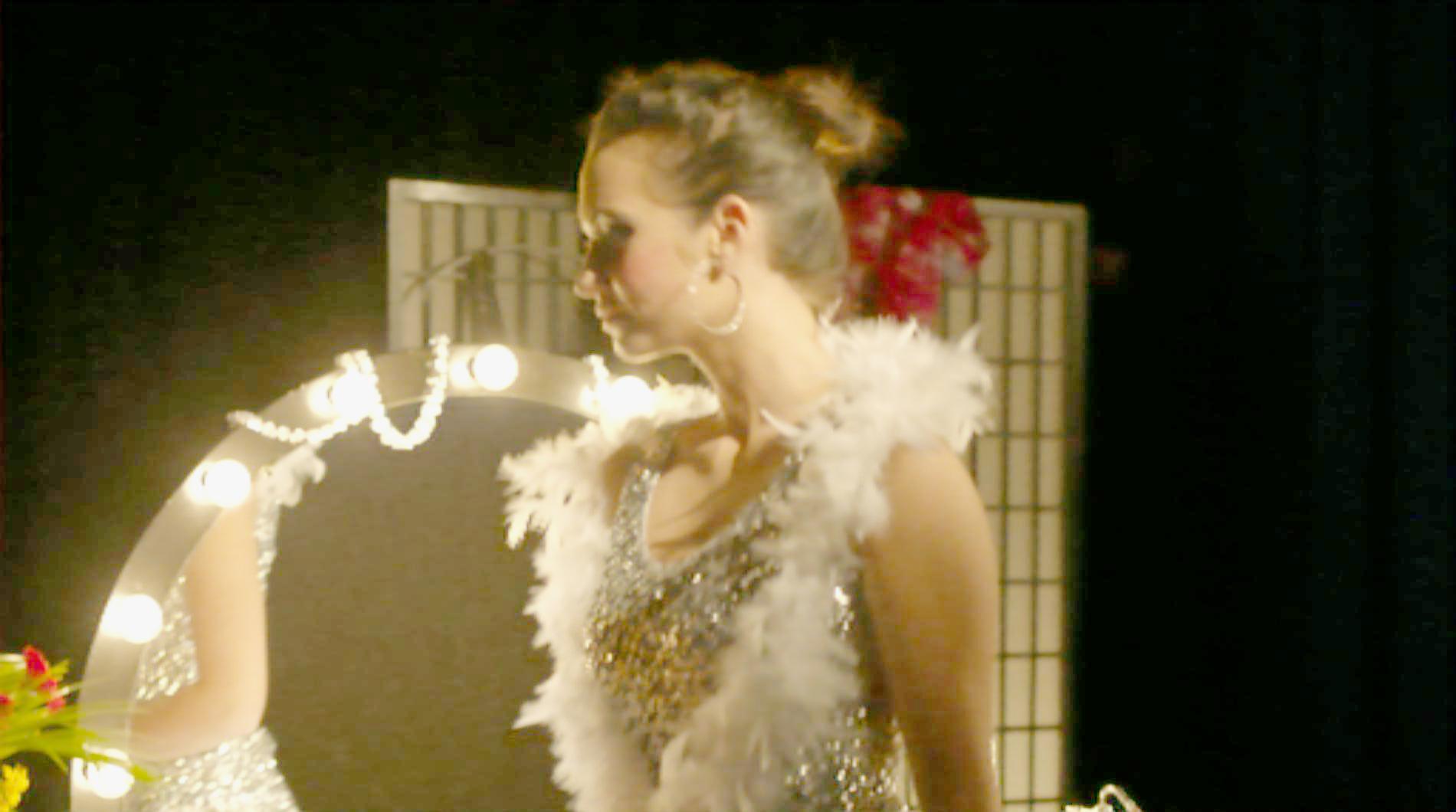}                & 
		\includegraphics[width = 0.2\textwidth]{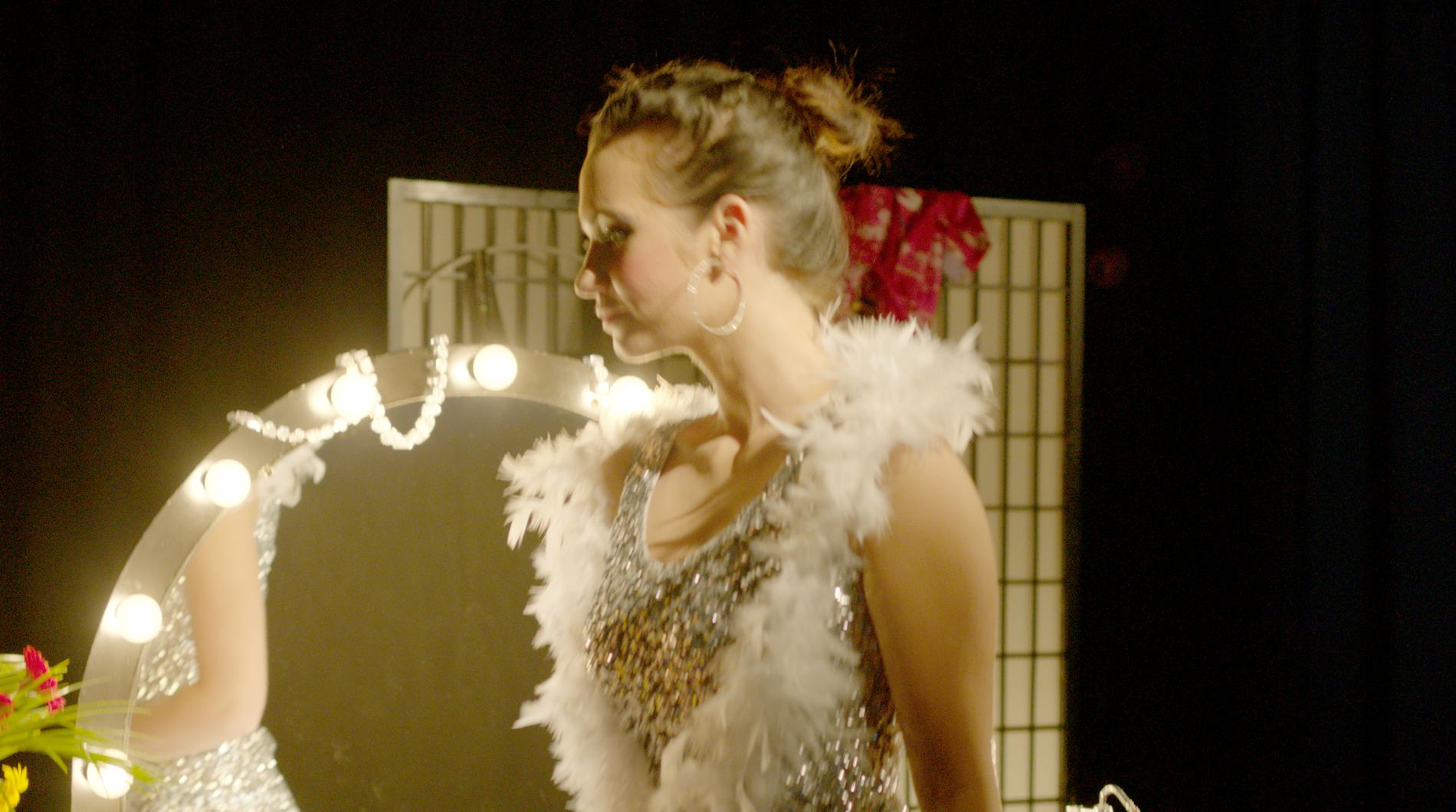}                 & 
		\includegraphics[width = 0.2\textwidth]{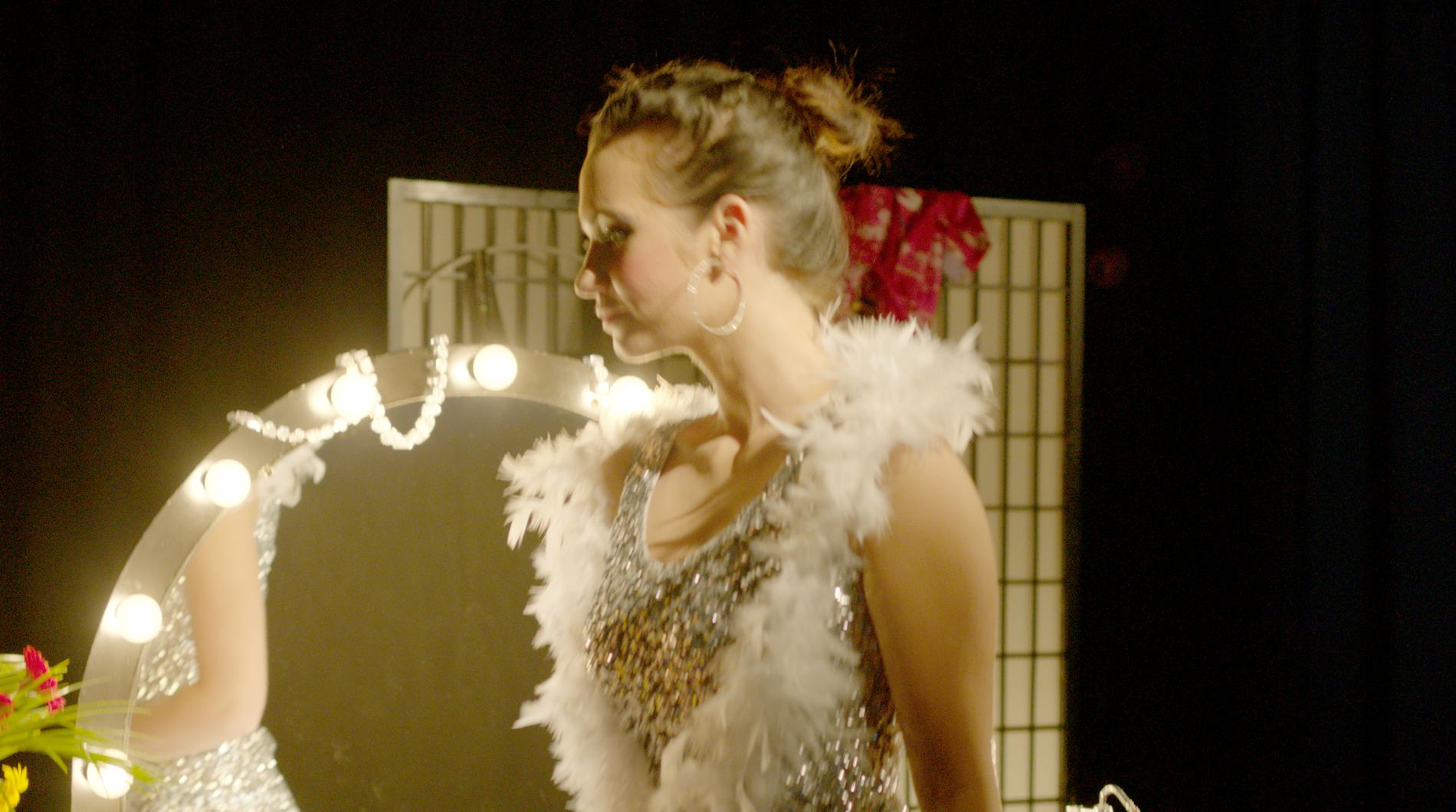}                     \\ 
		
		\includegraphics[width = 0.2\textwidth]{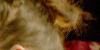}                 &
		\includegraphics[width = 0.2\textwidth]{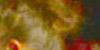}               &
		\includegraphics[width = 0.2\textwidth]{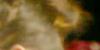}                &
		\includegraphics[width = 0.2\textwidth]{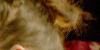}                 &
		\includegraphics[width = 0.2\textwidth]{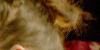}                       \\
		
		AHDR &
		U2fusion & 
		ADNet & 
		\textbf{Ours}&
		GT 
		\\

	\end{tabular}
	
	\caption{Our method obtains better visual quality and  more image details compared with other state-of-the-art methods in the NTIRE22 HDR datasets.}	
	\label{NTIRE}
	\vspace{-2mm}
\end{figure*}

\begin{figure*}[t]\scriptsize
	\tabcolsep 1pt
	\begin{tabular}{@{}cccccc@{}}
		\includegraphics[width = 0.16\textwidth]{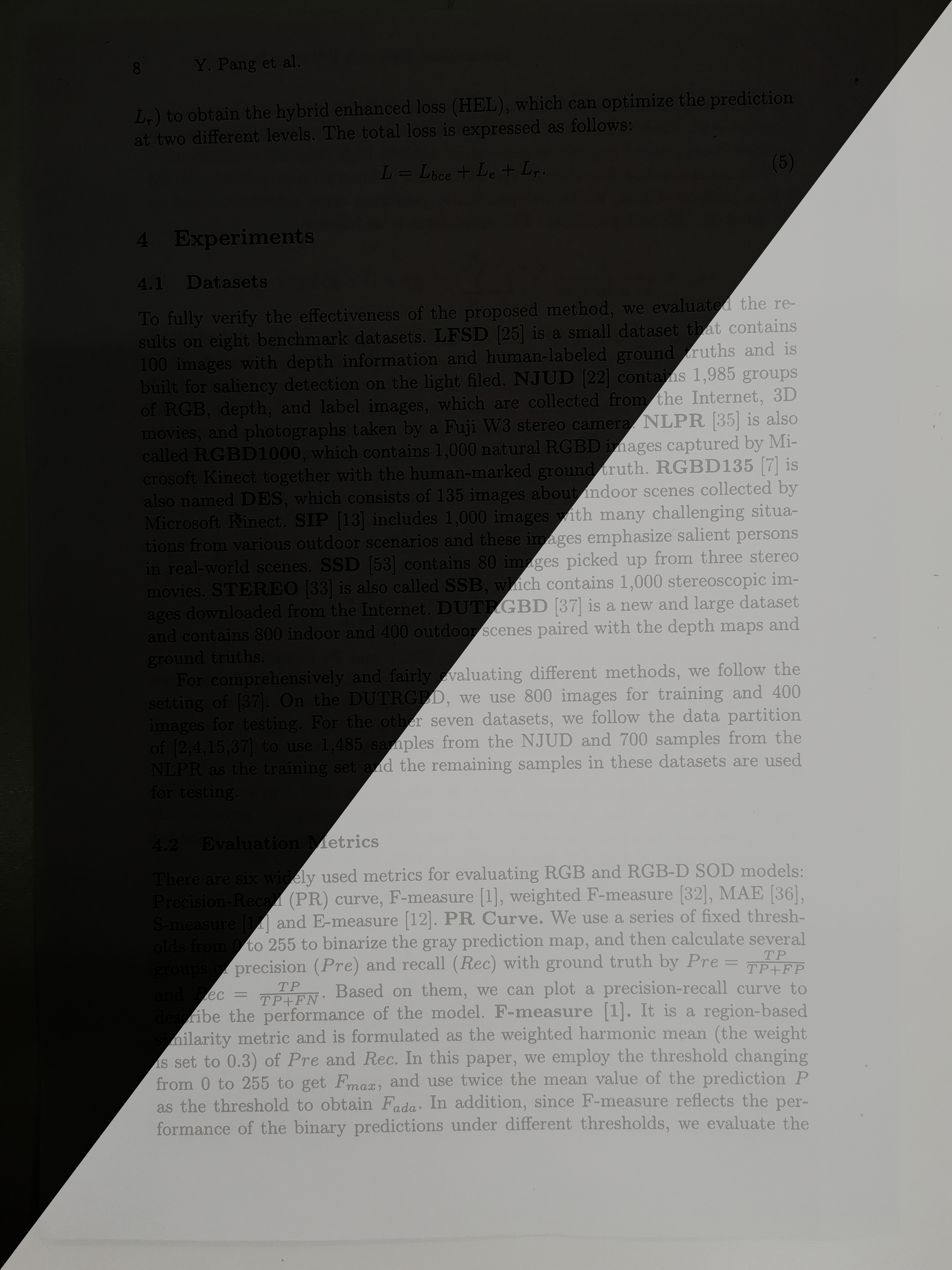}              &
		\includegraphics[width = 0.16\textwidth]{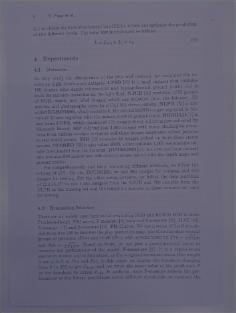}              &
		\includegraphics[width = 0.16\textwidth]{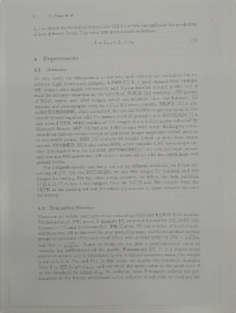}               &
		\includegraphics[width = 0.16\textwidth]{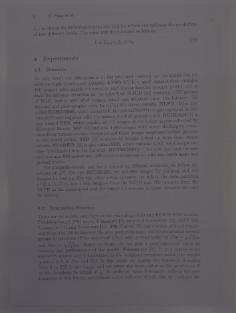}                 &
		\includegraphics[width = 0.16\textwidth]{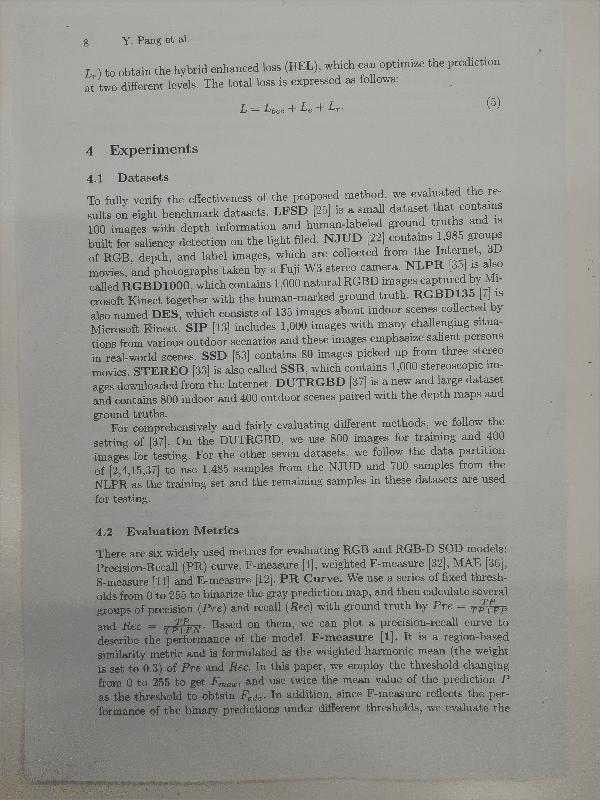}                 &
		\includegraphics[width = 0.16\textwidth]{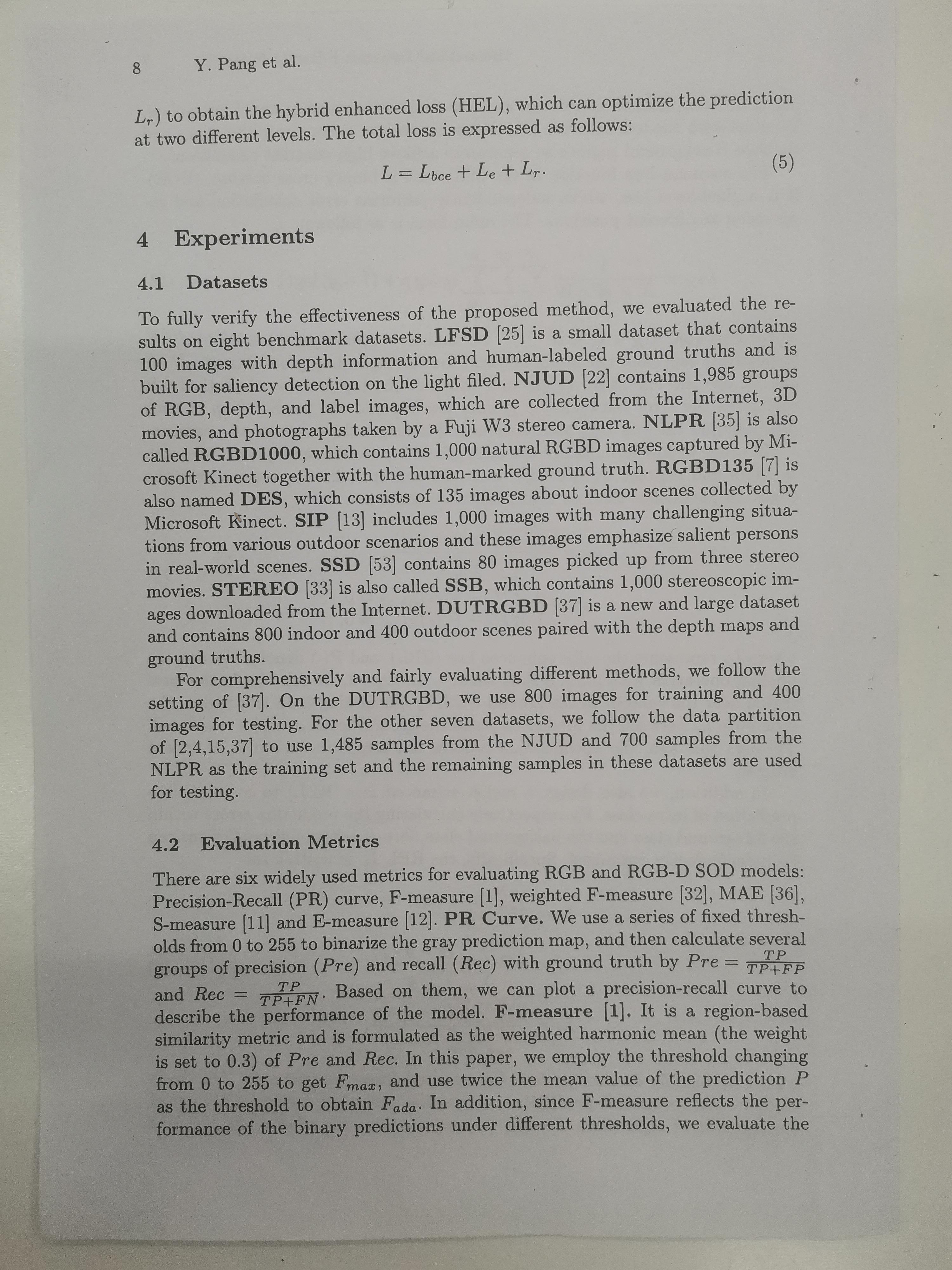}                       \\
		
		\includegraphics[width = 0.16\textwidth]{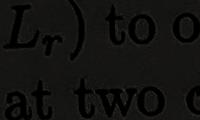}                 &
		\includegraphics[width = 0.16\textwidth]{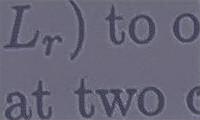}                 &
		\includegraphics[width = 0.16\textwidth]{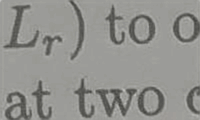}               &
		\includegraphics[width = 0.16\textwidth]{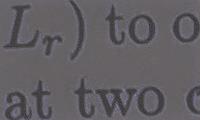}                &
		\includegraphics[width = 0.16\textwidth]{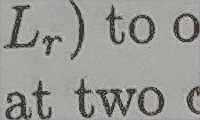}                &
		\includegraphics[width = 0.16\textwidth]{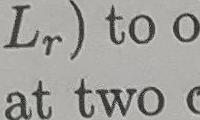}                       \\
		Input &
		AGAL& 
		U2fusion& 
		HALDER& 
		\textbf{Ours}& 
		GT\\
		
	\end{tabular}
	
	\caption{Our method obtains better visual quality and recovers more image details compared with other state-of-the-art methods in the MED dataset.}	
	\label{file}
\end{figure*}

\begin{figure*}[t]\scriptsize
	\tabcolsep 1pt
	\begin{tabular}{@{}cccc@{}}
		\includegraphics[width = 0.238\textwidth]{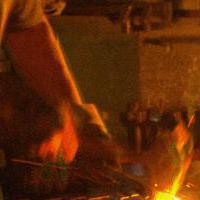}              &
		\includegraphics[width = 0.238\textwidth]{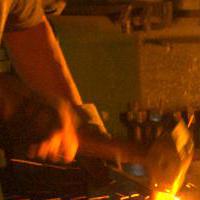}               &
		\includegraphics[width = 0.238\textwidth]{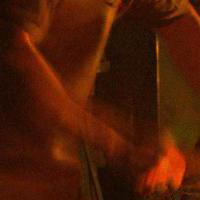}                &
		
		\includegraphics[width = 0.238\textwidth]{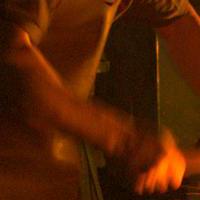}                       \\
		
		w/o teacher network &
		Ours & 
		w/o teacher network& 
		Ours\\
		
		

	\end{tabular}
	
	\caption{We select two groups of patches (local area of the image).  The architecture of the teacher-student network can provide a clearer texture for the images inferred by the student network.}	
	\label{shade_al}
	\vspace{-4mm}
\end{figure*}

\begin{figure*}[!h]\scriptsize
	\tabcolsep 1pt
	\begin{tabular}{@{}cccc@{}}
		
		\includegraphics[width = 0.238\textwidth]{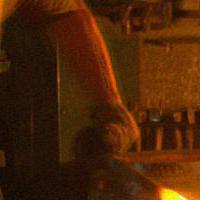}             & 
		\includegraphics[width = 0.238\textwidth]{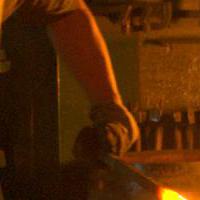}              &
		\includegraphics[width = 0.238\textwidth]{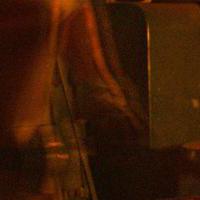}                 & 
		\includegraphics[width = 0.238\textwidth]{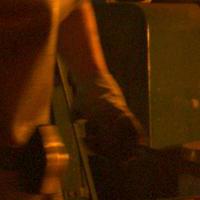}                     \\   
		
		w/o $\mathbf{L}_{lr}$ &
		Ours & 
		w/o $\mathbf{L}_{lr}$& 
		Ours\\

	\end{tabular}
	
	\caption{We select two groups of patches (local area of the image).  The architecture of the $\mathbf{L}_{lr}$ can provide a clearer texture for the images inferred by the student network.}	
	\label{shade_al2}
	\vspace{-0mm}
\end{figure*}

\noindent \textbf{Implementation details.} 
In the training phase, three input images with different exposure levels are divided into 1000 $\times$ 1000 resolution (bilinear interpolation or center cropping approach). 
The network is optimized by an AdamW optimizre with initial learning rate of 1e-4 and decay rate of 0.1, we set $\beta_1 = 0.9$, $\beta_2 = 0.999$ and $\varepsilon = 10^{-8}$ .  
Our network is implemented on PyTorch 1.7, Koinra, and Python 3.8. 
Our whole training process is implemented on RTX3090 GPU shader with 24G RAM.
The comparison algorithm is fine-tuned on all three datasets, and since these models do not have a customized lightweight design, the images are cropped to 512 $\times$ 512 during the training phase.
These networks are optimized by an AdamW optimizer with initial learning rate of 1e-3 and decay rate of 0.1, we set $\beta_1 = 0.9$, $\beta_2 = 0.999$ and $\varepsilon = 10^{-8}$. 
These comparison algorithms use a loss function that is an elaborate design for fine-tuning these three data sets.
Here, first, we select to use MEF-SSIM loss based on unreferenced metric method~\cite{DengZXGD21} to conduct semi-supervised training.
We used $I_k, k \in \{1,2,3\}$ to represent three input images with different exposure levels and used $I_p$ to represent the HDR image the model predicted.
\begin{equation}
	\begin{aligned}
		I_k &= \Vert{I_k-\mu_{I_k}\Vert}\cdot\frac{I_k-\mu_{I_k}}{\Vert I_k-\mu_{I_k} \Vert}+\mu_{I_k}\\
		&=c_k\cdot s_k + l_k,
	\end{aligned}
\end{equation}
where $\Vert{\cdot}\Vert$ is the $\mathcal{L}_2$ norm of pixel, $\mu_k$ is the mean value of $I_k$ and $\tilde{I_k}$ is the mean subtracted patch. The structure of the desired result ($\hat{s}$) is obtained by a weighted sum of structures of input patches. In addition, the desired contrast value $\hat{c}$  can be represented as follows:

\begin{equation}
	\begin{aligned}
		\hat{c}& = \text{max}\{c_k\},\\
		\bar{s}&=\frac{\sum_{k=1}^2w(\tilde I_k)s_k}{\sum_{k=1}^2 w(\tilde I_k)}~~\text{and}~~\hat{s} = \frac{\bar{s}}{{\Vert \bar{s} \Vert}},
		\\
		\hat{I} &= \hat{c}\cdot\hat{s},
	\end{aligned}
\end{equation}
\noindent
where the $\hat{I}$ stands for the result. 
The final image quality score for pixel $p$ is calculated using the SSIM framework:
\begin{equation}
	\text{Score}(p) = \frac{2\sigma_{\hat{I}_p}+C}{\sigma_{\hat{I}}^2+\sigma_{I_p}^2+C},
\end{equation}
\noindent
where $\sigma_{\hat{I}}^2$ is variance and $\sigma_{\hat{I}_p}$ is covariance between $\hat{I}$ and $I_p$ and the final loss is calculated as follows:
\begin{equation}
	\mathcal{L}_{t} = 1-\frac{1}{N}\sum_{p \in P}\text{Score}(p),
\end{equation}
\noindent
where $N$ is the total number of pixels in the input image and $P$ are the set of all pixels in the input image. In addition, we used traditional $\mathcal{L}_{1}$ to train for paired datasets. Therefore, our total loss can be expressed as follows:
\begin{equation}
	L_{\text{total}}=\lambda_1{L_{\text{paired}}+\lambda_2{L_{\text{unpaired}}}},
\end{equation}
\noindent
where $\lambda_1$ and $\lambda_2$ represent the weight of paired and unpaired data loss and we set the $\lambda_1$ 0.9 and the $\lambda_2$ 0.1.
In addition, since AHDR~\cite{yan2019attention} and ADNet\cite{liu2021adnet} cannot run MED directly on a single GPU, for this reason, we sub-patch the images into the model before stitching.

\noindent \textbf{Algorithm comparison.}
We compared our method with HALDER~\cite{9428192}, AGAL~\cite{liu2022attention}, AHDR~\cite{yan2019attention}, U2fusion~\cite{xu2020u2fusion}, ADNet~\cite{liu2021adnet} on SICE~\cite{cai2018learning} dataset and the NTIRE workshop22 Multi-frame HDR dataset. 
%
Figure~\ref{SICE} shows two examples from the SICE dataset,  Figure~\ref{NTIRE} shows two examples from NTIRE workshop22 Multi-frame HDR dataset and Figure~\ref{file} shows an example from MED. 
Through the demonstration of the samples, we find that U2fusion can generate artifacts when fusing images, this is because the unsupervised method is not stable. 
AGAL's attention mechanism is more focused on fusing the textures of the images, while the colors are generally distorted. 
AdNet over-smoothed the texture.
As shown in Table~\ref{table:headings}, we evaluate the performance of the algorithm in terms of quantitative aspects, and in general, our method reaches the best in terms of speed and accuracy.

\noindent \textbf{Ablation study.}
We conduct some ablation experiments to evaluate the effectiveness of the modules of our method.
1) \textcolor{blue}{Effectiveness of teacher-student network}.
We remove the teacher network and use only the student network to fuse the multi-exposure images.
These networks are optimized by an AdamW optimizer with an initial learning rate of 1e-5 and a decay rate of 0.1, we set $\beta_1 = 0.9$, $\beta_2 = 0.999$ and $\varepsilon = 10^{-8}$. 
For the loss function term, we use $\mathcal{L}_1$ and perceptual loss.
As shown in Figure~\ref{shade_al}, our method provides clearer textures when reconstructing the local information of the image.
2) \textcolor{blue}{Effectiveness of $\mathcal{L}_{lr}$}.
We remove the loss function term $\mathcal{L}_{lr}$ with the correlation between patches.
As shown in Figure~\ref{shade_al}, our method provides clearer textures when reconstructing the local information of the image.
3) \textcolor{blue}{Effectiveness of 3D LUT grid}.
Our proposed implicit neural network can be run in edit mode to generate a 3D LUT grid of arbitrary size, and here we evaluate the role of different scale grids on the SICE dataset (see Table~\ref{unpaired_num}).
We observe an optimal trade-off between speed and accuracy when the grid size reaches 32.
In addition to these, we quantitatively evaluate the ablation experiments on the SICE and MED dataset (see Table~\ref{ad}).

\begin{table}[!t]
	\centering
	\caption{Quantified results of ablation experiments for network components. TN denotes Teacher Network.}
	\resizebox{0.48\textwidth}{!}{
		\begin{tabular}{ccccccc}
			\toprule
			\multirow{2}{*}{Datasets} & \multicolumn{2}{c}{w/o TN} & \multicolumn{2}{c}{w/o $\mathcal{L}_{lr}$} & \multicolumn{2}{c}{Ours}  \\
			\cmidrule(lr){2-3}\cmidrule(lr){4-5}\cmidrule(lr){6-7}
			& PSNR & SSIM & PSNR  & SSIM & PSNR & SSIM  \\
			\midrule
			SICE & 21.22 & 0.72 & 23.6 & 0.80 & \textbf{23.8} & \textbf{0.81}  \\
			MED & 25.69 & 0.89 & 27.2 & 0.94 & \textbf{29.78} & \textbf{0.97}   \\
			\bottomrule
	\end{tabular}}
	\label{ad}
\end{table}

\begin{table}
	\begin{center}
		\caption{The effect of different sizes of 3D LUT grids on reconstructed images in student networks.}
		\label{unpaired_num}
		\setlength{\tabcolsep}{3.1mm}{
			\begin{tabular}{cccccc}
				\toprule	
				& 8    &  16   & 32    & 64     & 128   \\ \midrule
				PSNR & 15.4 &  22.1 & 23.8  & 24.1   & 24.2\\	
				SSIM & 0.61 &  0.77 & 0.81  & 0.85  & 0.85\\			
				\bottomrule
		\end{tabular}}
		\vspace{-4mm}
	\end{center}
\end{table}

\begin{figure}[h]\scriptsize
	\begin{center}
		\begin{tabular}{@{}c@{}}
			\includegraphics[width = 0.35\textwidth]{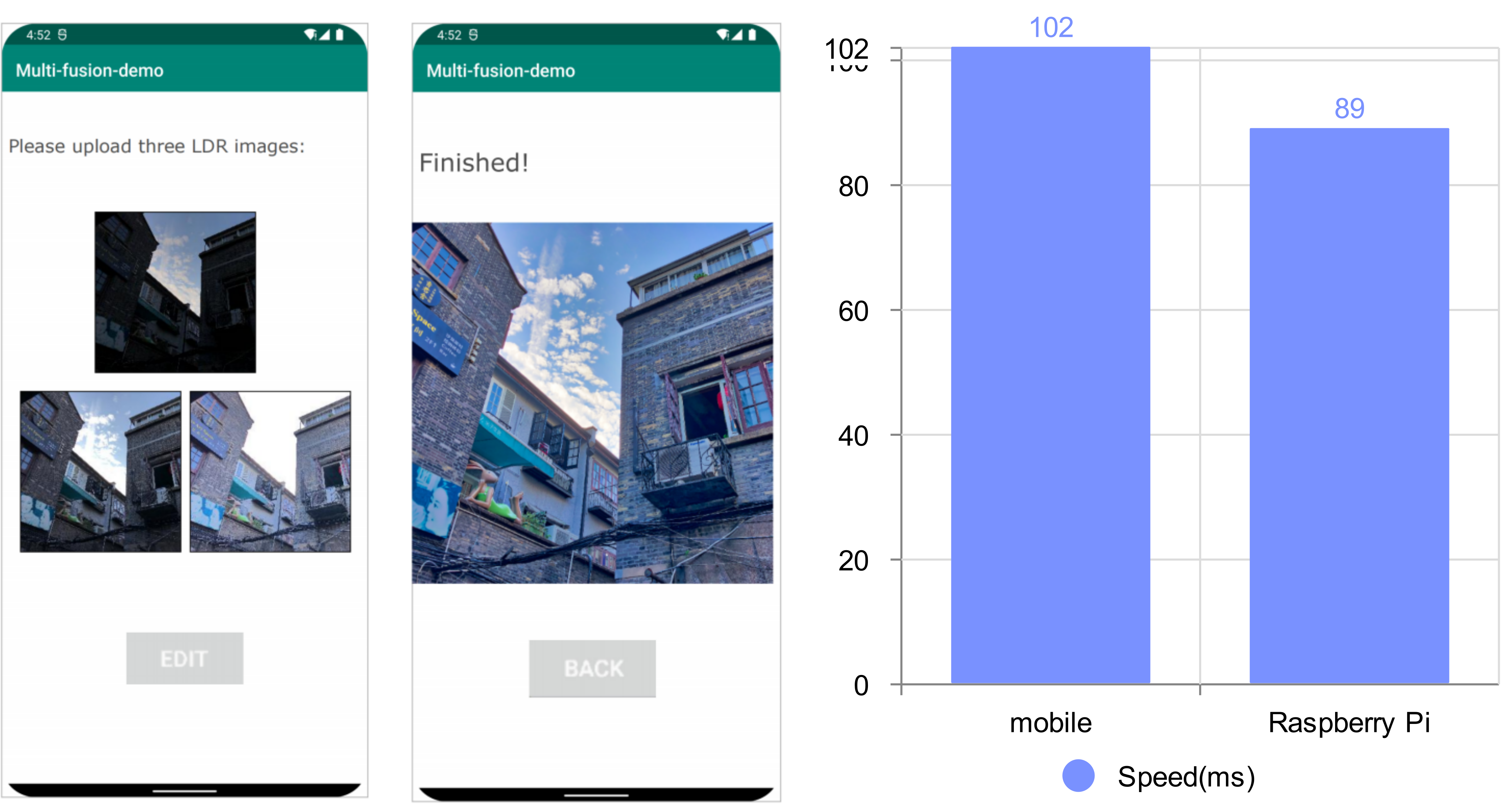}                                 \\
			
		\end{tabular}
		
		\caption{Results of the program in the Android studio environment.}	
		\label{mobile}
		\vspace{-4mm}
	\end{center}
\end{figure}

\noindent \textbf{Deployment of mobile devices.}
We also deploy our proposed network on Android virtual mobile (102ms) and Raspberry PI (89ms) for the real-time image processing requirements of mobile devices. 
In addition, Table~\ref{table:headings} shows the number of parameters for our proposed method and others. 
We used the API interface provided by PyTorch mobile to embed the quantized model into the Android development program and experimented with the Pixel 4 XL API 32 virtual machine in Android Studio.
Some of the experimental results are shown in Figure~\ref{mobile}.
\begin{figure}[h]\scriptsize
	\tabcolsep 1pt
	\begin{tabular}{@{}cccc@{}}
		\includegraphics[width = 0.115\textwidth]{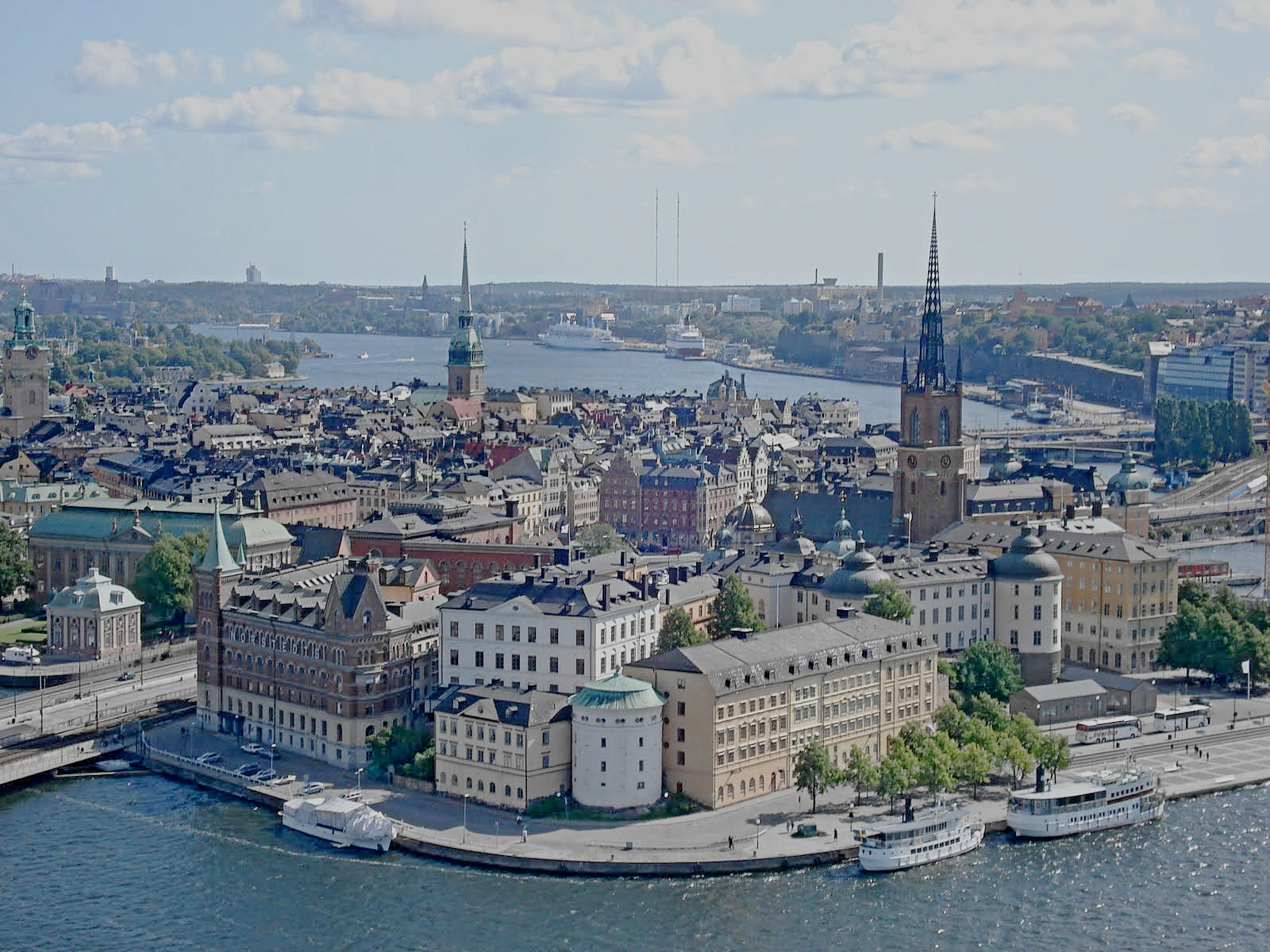}              &
		\includegraphics[width = 0.115\textwidth]{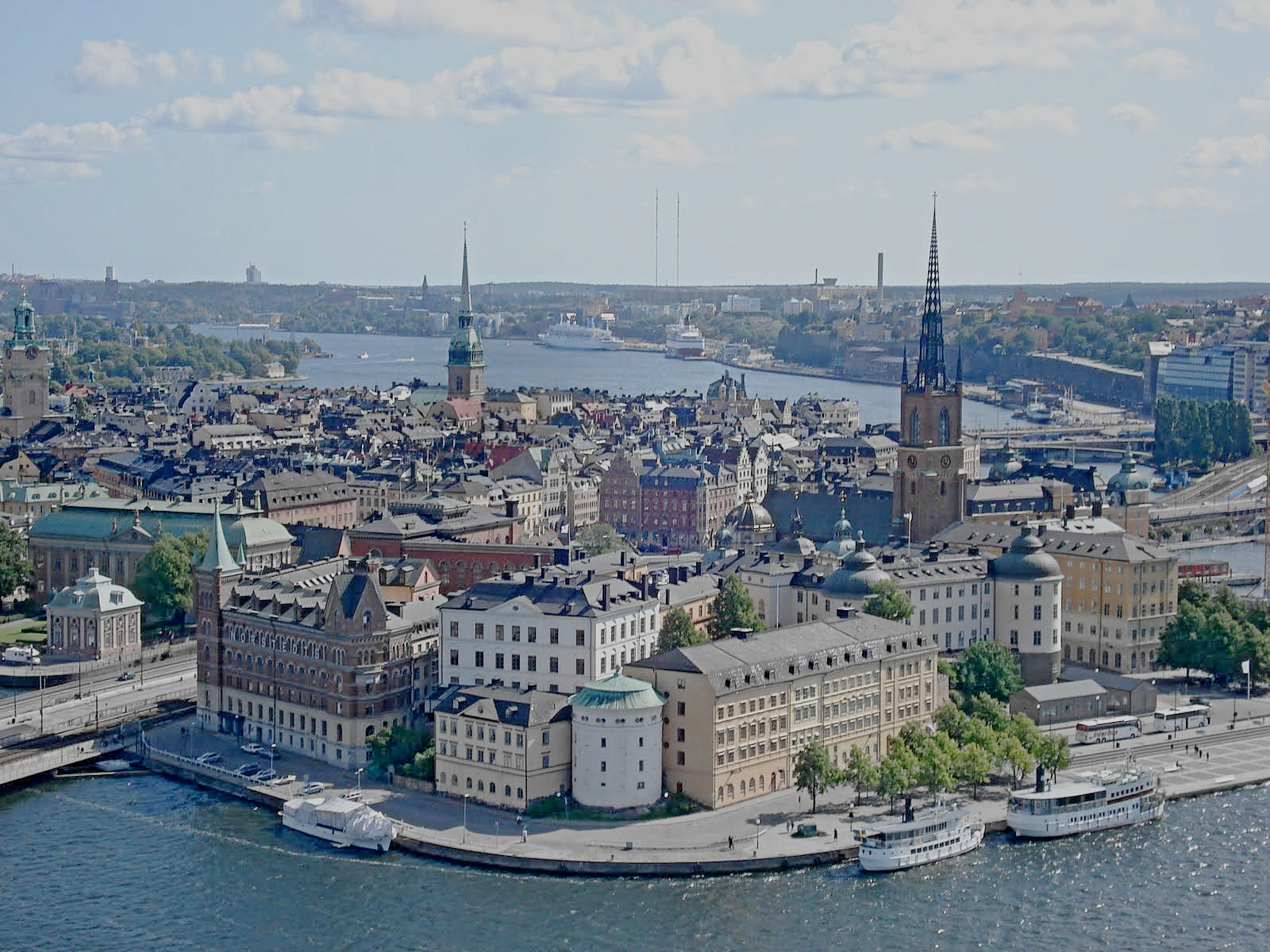}                    &
		\includegraphics[width = 0.115\textwidth]{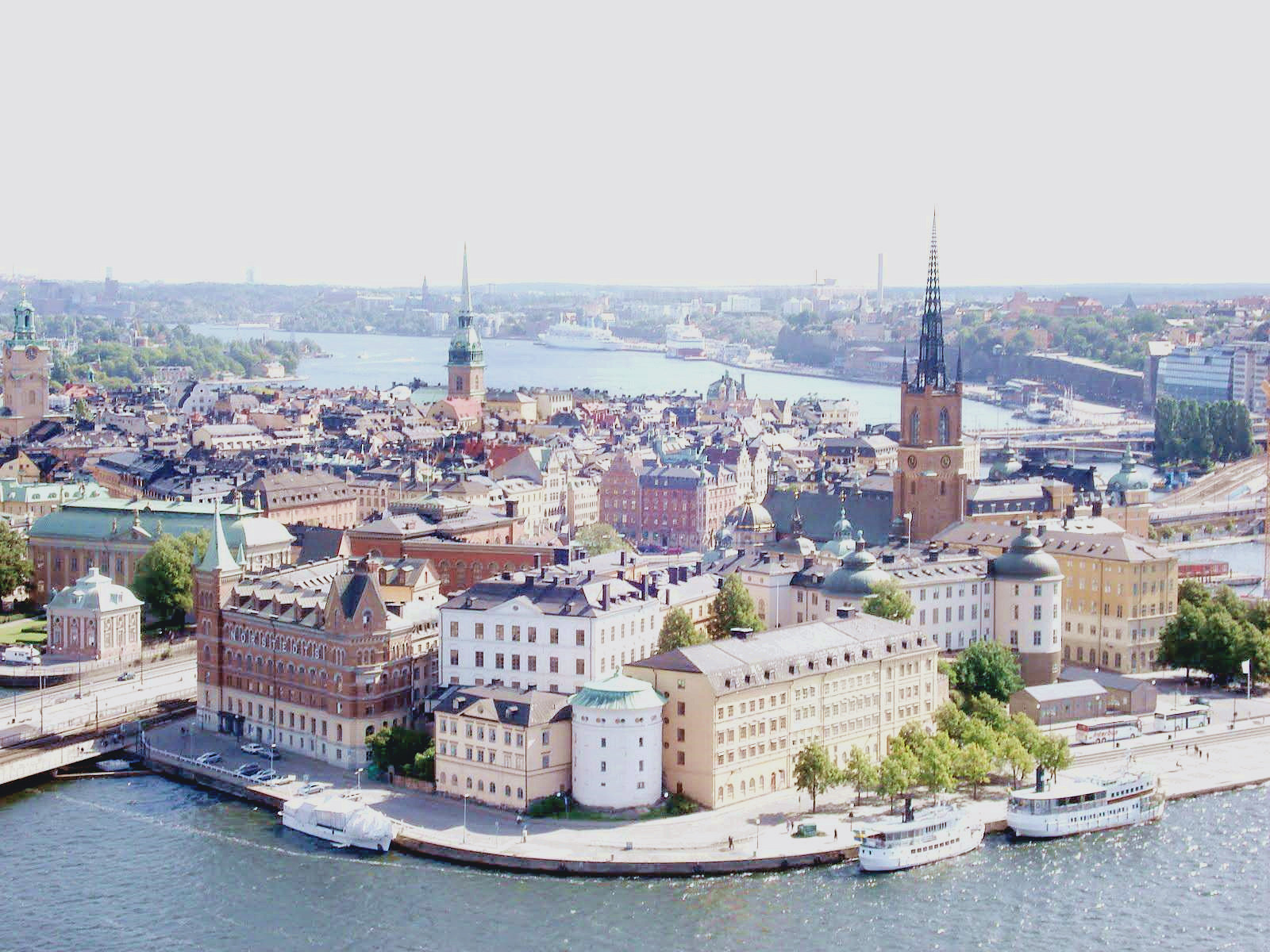}              &
		\includegraphics[width = 0.115\textwidth]{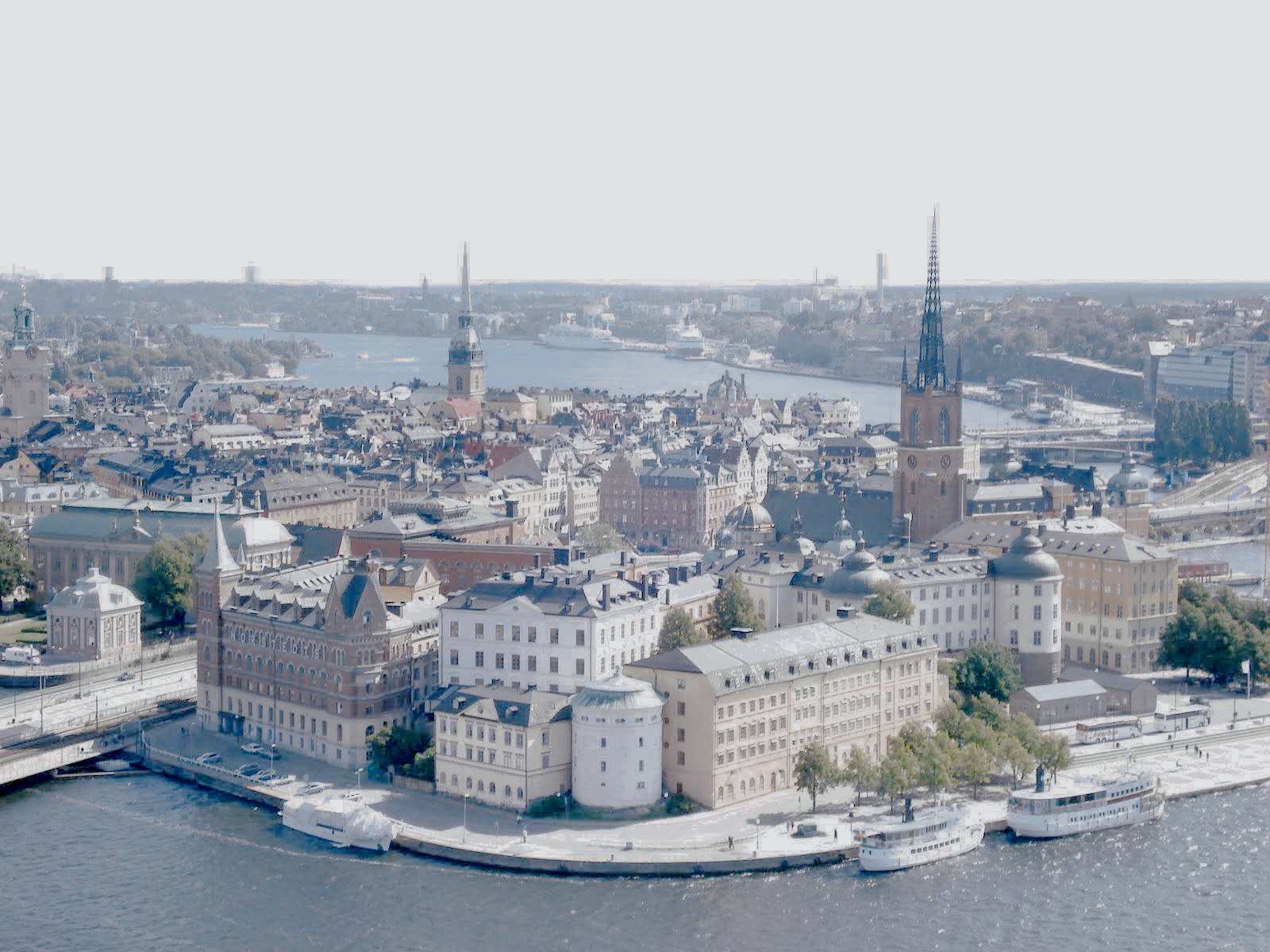}                               \\
		\includegraphics[width = 0.115\textwidth]{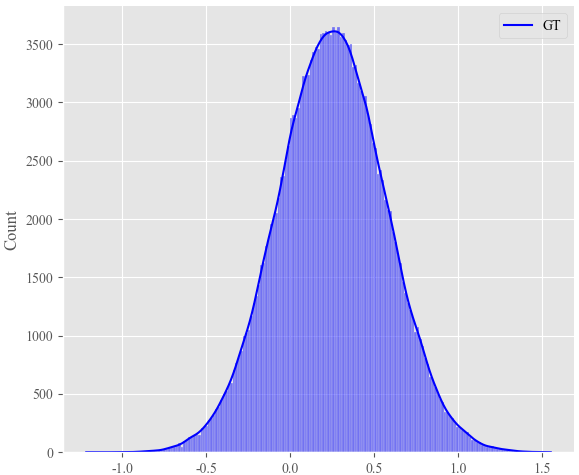}              &
		\includegraphics[width = 0.115\textwidth]{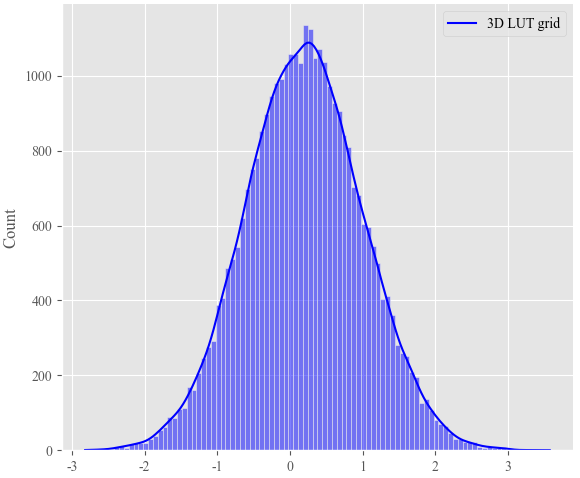}                    &
		\includegraphics[width = 0.115\textwidth]{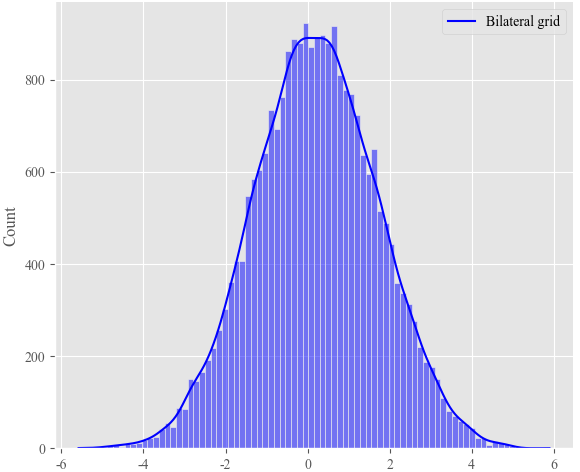}              &
		\includegraphics[width = 0.115\textwidth]{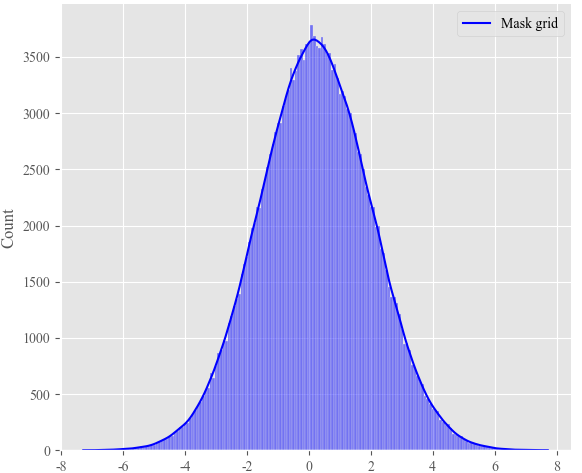}                               \\
		
		(a) GT &
		(b) 3D LUT &
		(c) HDRNet &
		(d) Pyramid \\
		
	\end{tabular}
	
	\caption{This figure shows the reconstruction results of the networks and the data distribution of the corresponding grids.}	
	\label{d}
	\vspace{-6mm}
\end{figure}

\begin{figure}[h]\scriptsize
	\tabcolsep 1pt
	\begin{tabular}{@{}ccc@{}}
		\includegraphics[width = 0.155\textwidth]{d/5.jpg}              &
		\includegraphics[width = 0.155\textwidth]{d/lut3d.jpg}              &
		\includegraphics[width = 0.155\textwidth]{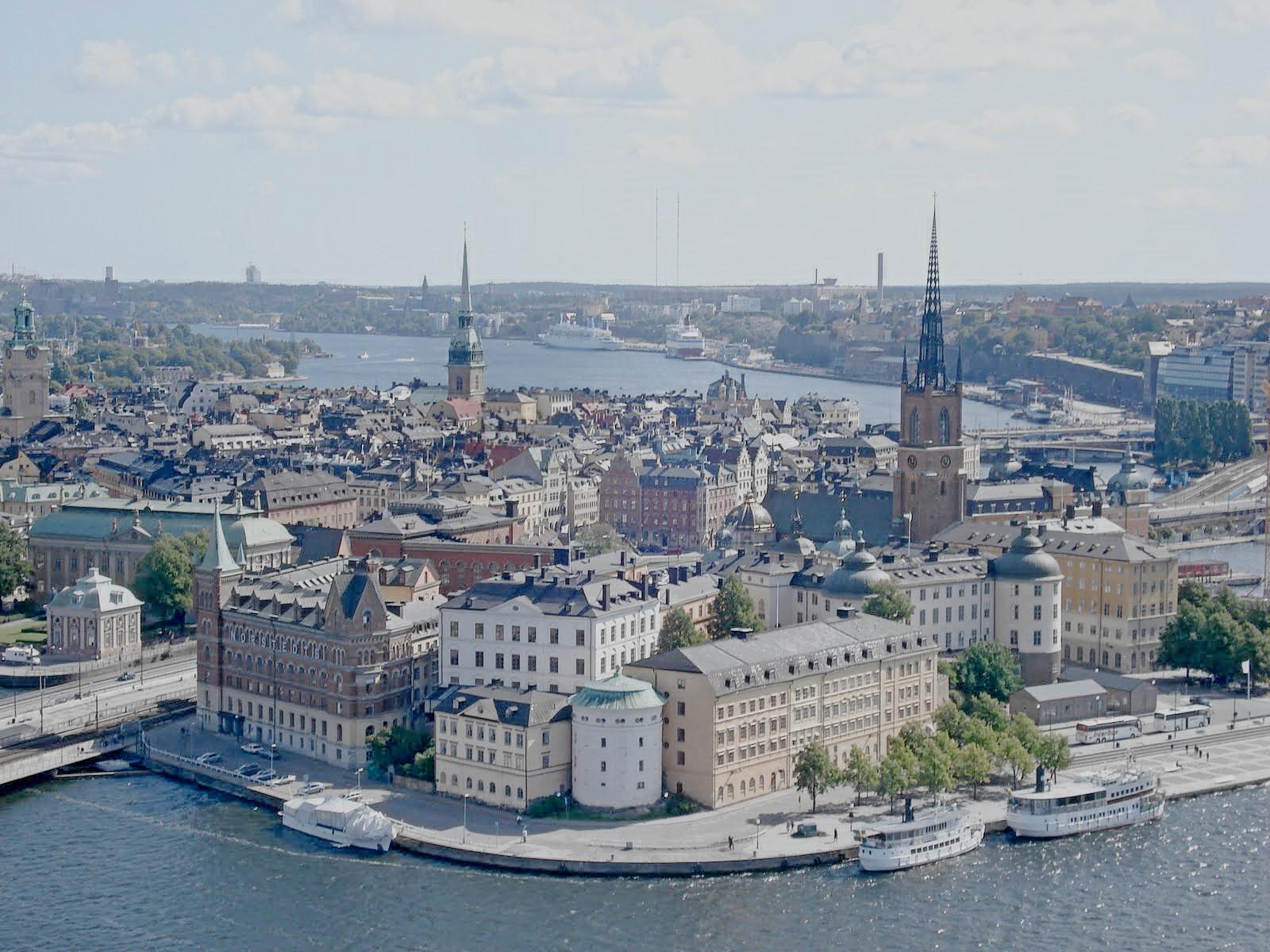}                    \\
		
		(a) GT &
		(b) 3D LUT &
		(c) RI \\
	\end{tabular}
	
	\caption{This figure shows the results of using randomly initialized 3D LUTs (RI) instead of the basic 3D LUTs.}	
	\label{d2}
	\vspace{-6mm}
\end{figure}

\section{Discussion}
%
%
\textit{Why we chose 3D LUT to model multi-exposure image fusion with UHD resolution rather than bilateral learning or pyramids?}
We conduct two comparison experiments where they (3D LUT, HDRNet, Laplace Pyramid) are performed on a degraded UHD image to obtain a high-quality UHD image.
These networks are fine-tuned on the MIT-Adobe FiveK dataset.
They are optimized by an AdamW optimizer with an initial learning rate of 1e-3 and a decay rate of 0.1, we set $\beta_1 = 0.9$, $\beta_2 = 0.999$ and $\varepsilon = 10^{-8}$. 
We visualize the data distribution of the 3D LUT grid (32 $\times$ 32 $\times$ 32), bilateral grid (12 $\times$ 16 $\times$ 16 $\times$ 8), and MASK grid (3 $\times$ 256 $\times$ 256) (see Figure~\ref{d}).
In general, the data distribution of the grids generated by these networks is close; the mean value of the distribution is almost always around 0.2.
However, the variance of each grid is different, and the variance of 3D LUT is the smallest.
We assume that this is related to a priori knowledge and remove the basic 3D LUTs and replace them with a fixed distribution grid.
The results demonstrate that the variance of the grid becomes larger ([-3, 3] $\rightarrow$ [-5, 5]) and the reconstruction of the image is degraded (see Figure~\ref{d2}).
%

\section{Conclusion}
In this paper, we propose a teacher-student network based on a 3D LUT to achieve real-time (33fps) UHD multi-exposure image fusion on a single GPU.
In addition, we develop an editable pattern to obtain HDR images of different quality to adapt to different scenes.
We discuss the reasons for using 3D LUTs by trading off the two aspects of speed (ms) and accuracy (PSNR).
Extensive experimental results validate the effectiveness of our method. 
Note that the reconstruction effect is not significant for 3D LUTs with grid scales larger than 64.

\bibliographystyle{ieee_fullname}
\bibliography{egbib}

\end{document}